\documentclass[10pt,twocolumn,letterpaper]{article}
\usepackage[accsupp]{axessibility}  %
\usepackage{iccv}
\usepackage{times}
\usepackage{epsfig}
\usepackage{graphicx}
\usepackage{amsmath}
\usepackage{amssymb}

\usepackage{caption}
\usepackage{subcaption}

\usepackage{booktabs}
\usepackage{multirow}
\usepackage{array}
\usepackage{tabularx}

\usepackage{mathrsfs}
\usepackage{bm}
\usepackage{float}
\usepackage{wrapfig}

\usepackage[pagebackref=true,breaklinks=true,letterpaper=true,colorlinks,bookmarks=false]{hyperref}

\newcommand{\myparagraph}[1]{\vspace{1pt}\noindent{\bf{#1}}}
\iccvfinalcopy %

\usepackage[symbol]{footmisc}

\def\iccvPaperID{9517} %

\ificcvfinal\fi

\begin{document}

\vspace{-35pt}
\title{\texttt{ProbVLM}: Probabilistic Adapter for Frozen Vision-Language Models %
}

\vspace{-35pt}

\author{\vspace{0.5em}
\setlength\tabcolsep{0.5em}
\begin{tabular}{cccc} 
Uddeshya Upadhyay$^{*,1}$ & Shyamgopal Karthik$^{*,1}$ & Massimiliano Mancini$^{2}$ & Zeynep Akata$^{1,3}$ \tabularnewline
\end{tabular}
\\
\renewcommand{\arraystretch}{0.5}
\begin{tabular}{cccc} 
    $^1$University of T\"{ubingen} & $^2$University of Trento & $^3$MPI for Intelligent Systems
\end{tabular}
}

\maketitle

\footnotetext[1]{Authors contributed equally.}
\thispagestyle{empty}

\begin{abstract}
Large-scale vision-language models (VLMs) like CLIP successfully find correspondences between images and text. Through the standard deterministic mapping process, an image or a text sample is mapped to a single vector in the embedding space. This is problematic: as multiple samples (images or text) can abstract the same concept in the physical world, deterministic embeddings do not reflect the inherent ambiguity in the embedding space. We propose \texttt{ProbVLM}, a probabilistic adapter that estimates probability distributions for the embeddings of pre-trained VLMs via inter/intra-modal alignment in a post-hoc manner without needing large-scale datasets or computing. On four challenging datasets, i.e., COCO, Flickr, CUB, and Oxford-flowers, we estimate the multi-modal embedding uncertainties for two VLMs, i.e., CLIP and BLIP, quantify the calibration of embedding uncertainties in retrieval tasks and show that \texttt{ProbVLM} outperforms other methods. Furthermore, we propose active learning and model selection as two real-world downstream tasks for VLMs and show that the estimated uncertainty aids both tasks. Lastly, we present a novel technique for visualizing the embedding distributions using a large-scale pre-trained latent diffusion model. Code is available at \url{https://github.com/ExplainableML/ProbVLM}. 
\end{abstract}

\section{Introduction}
\label{sec:intro}

\begin{figure}
    \centering
    \includegraphics[width=0.47\textwidth]{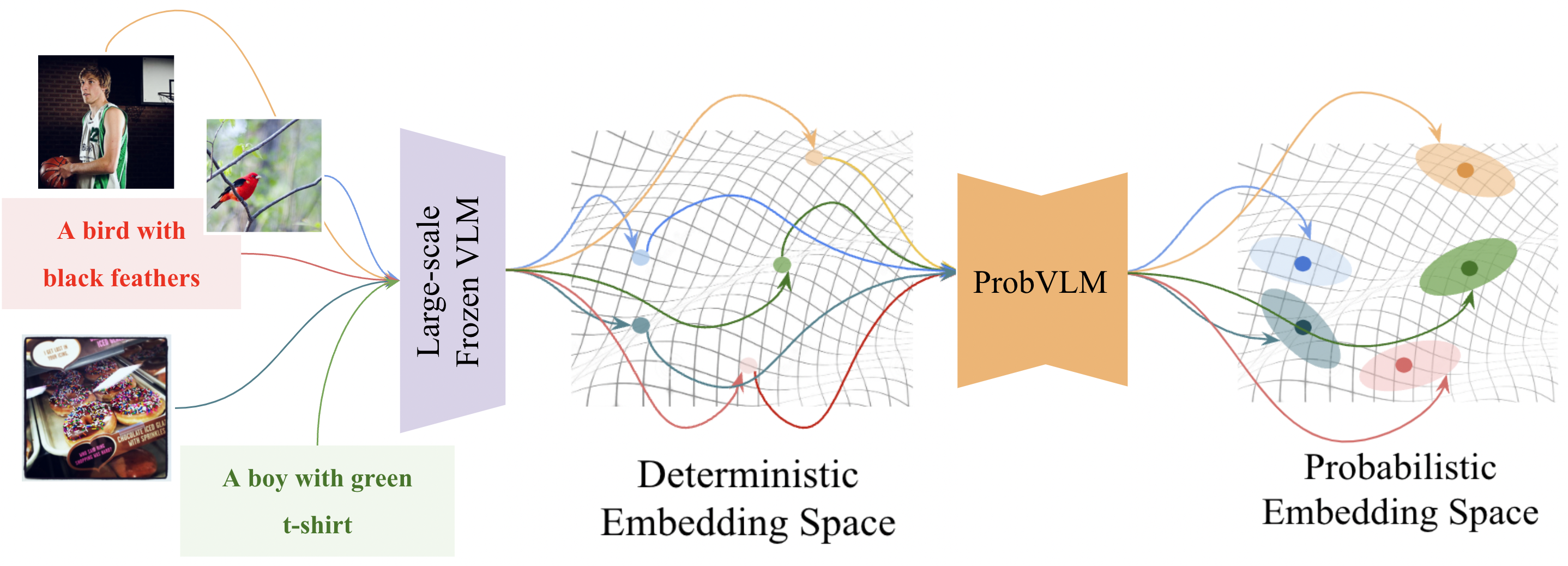}
    \caption{
    We provide probabilistic embeddings for deterministic pre-trained  vision-language models that are \textit{frozen}. By capturing the ambiguity inherently present in the inputs, we obtain well-calibrated uncertainty estimates.
    }
    \vspace{-10pt}
    \label{fig:teaser}
\end{figure}
Recently, large vision-language models (VLMs)~\cite{clip,slip,blip,flava,flamingo,align} have become exceedingly popular due to their ability to align images and text. These models such as CLIP~\cite{clip} and BLIP~\cite{blip}
are trained on large-scale datasets such as LAION-400M~\cite{laion} and YFCC-100M~\cite{yfcc} and have shown strong performance when evaluated in a zero-shot fashion (i.e without requiring fine-tuning on specific datasets) for a variety of downstream tasks. One of the most popular applications of VLMs is cross-modal retrieval~\cite{wang2017adversarial,wang2016comprehensive} i.e retrieving images (text) for a queried text (images). However, image-to-text matching (and vice-versa) is fundamentally ill-posed due to the inherent ambiguity in either modality~\cite{yang2021probabilistic}, i.e. the same caption (or image) can be valid for multiple images (or captions). Therefore, it becomes essential to model the ambiguity inherently present in the various modalities, and combinations thereof. 

Instead of mapping inputs to embeddings, probabilistic embedding methods~\cite{hib,pcme} learn to map input samples to distributions. 
This is achieved by parameterizing the distributions of the embeddings and training a deep neural network to maximize its likelihood. Although they model ambiguities in the embedding space, 
such probabilistic models require training deep networks from scratch. This requires access to the large-scale datasets and the computational resources of the recent VLMs~\cite{clip,align,slip,flava,blip}. 

We propose \texttt{ProbVLM}, a post-hoc probabilistic adapter, the first method to convert the deterministic embeddings provided by a \textit{frozen} large-scale vision-language models into probabilistic ones, as shown in Figure~\ref{fig:teaser}. This enables us to efficiently retain the benefits of large-scale pre-training while learning distributions that model the inherent ambiguities in the different modalities. 
Our \texttt{ProbVLM} models the embedding distribution as a heteroscedastic probability distribution and is trained using a combination of intra-modal and cross-modal alignment objectives and provides well-calibrated uncertainty estimates, useful for several tasks.

We demonstrate on two large vision-language datasets, i.e., COCO~\cite{mscoco} and Flickr~\cite{flickr30k}, and on two fine-grained image datasets, i.e., CUB~\cite{cub} and Oxford-Flowers~\cite{flo} with sentences from \cite{cubcaptions}, that \texttt{ProbVLM} learns calibrated uncertainties without requiring large-scale models to be trained from scratch. This sharply contrasts previous works on probabilistic embeddings~\cite{hib,pcme} that train new models from scratch.
We perform a series of analyses to understand the impact of the training objective and to study the properties of the resulting uncertainties. 
Furthermore, we demonstrate that
our uncertainty estimates can be used to select the optimal model from a set of finetuned vision-language models on an unlabeled target dataset. They can also be used to choose the most suitable samples for fine-tuning the model in an active learning setup. 
Finally, with the help of a pretrained latent diffusion model~\cite{ldm}, i.e., \textit{Stable Diffusion}, we decode sampled embeddings from predicted distribution to visualize the predicted embedding distributions. We show that the predicted embedding distributions indeed capture meaningful modes of variation, that may be interpretable.

\begin{figure*}[!t]
    \centering
    \includegraphics[width=0.98\textwidth]{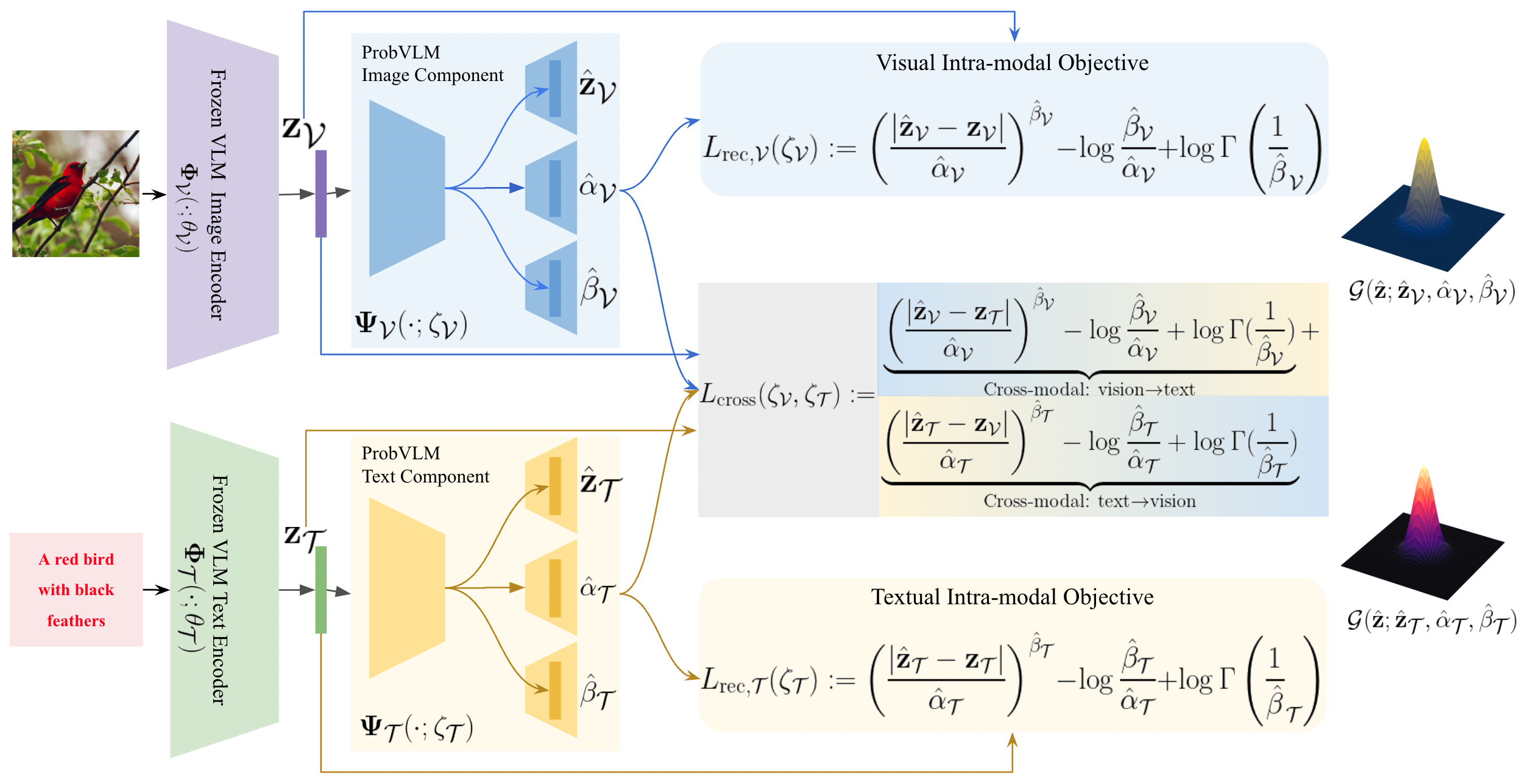}
    \vspace{-6pt}
    \caption{Proposed framework (\texttt{ProbVLM}) takes an existing vision-language model and introduces a probabilistic adapter over the image and text encoders. These adapters predict the parameters of a parameterized distribution for a given embedding. Models are trained by minimizing an objective consisting of intra/cross-modal supervision as detailed in Section~\ref{sec:method}.}
    \vspace{-10pt}
    \label{fig:arch}
\end{figure*}

\section{Related Work}
\myparagraph{Vision-Language Models.} Such models~\cite{clip,slip,flava,flamingo,blip,vilbert,blipv2,filip,coca,simvlm} have become ubiquitous in recent times due to their various applications in image classification~\cite{zhou2022conditional,clip-adapter,coop,ming2022delving}, cross-modal retrieval~\cite{bain2022clip}, as well as open-vocabulary semantic segmentation~\cite{ghiasi2021open,xu2021simple}. The most notable among these is CLIP~\cite{clip}, which consists of an image and text encoder trained on 400M image-text pairs with a contrastive objective~\cite{nce,infonce}. 
As a result, the model is able to project images and text to a shared embedding space. 
In this paper, we focus on using the shared embedding space for the task of cross-modal retrieval~\cite{flickr30k,mscoco}.
 Recent works have predominantly relied on large-scale pre-training~\cite{clip,slip,flava,flamingo,vinvl,laion,laion5b} to project images and text to the same metric space. However, it is essential to note that all of these vision-language models~\cite{clip,slip,blip,flava,flamingo} provide deterministic mappings that do not model the inherent ambiguity in the inputs. %
 In this work, we turn a deterministic model (i.e.,  CLIP) into a probabilistic one, without the need of a large-scale dataset.

\myparagraph{Probabilistic Embeddings.} These methods~\cite{hib,pcme,lidifferentiable} provide an elegant solution to estimate the ambiguity present in the inputs~\cite{kirchhof2023arxiv}. The key idea here is to map inputs to probability distributions in the embedding space, as opposed to point estimates, thereby modeling the inherent ambiguity present in the input. In the context of cross-modal retrieval, this was done by optimizing a probabilistic analog of the contrastive objective to learn distributions for the image and text inputs~\cite{pcme}. Other works have further improved the performance~\cite{lidifferentiable,park2022probabilistic,ji2022map}, extended this formulation to achieve compositional retrieval~\cite{neculai2022probabilistic}, and have applied it to other tasks such as video retrieval~\cite{park2022probabilistic,fang2023uatvr} and tasks like pose estimation~\cite{sun2020view}. However, most of these works focus on training a model from scratch, thereby not leveraging the power of the pre-trained models that are widely present. The notable exception to this is Probabilistic Face Embedding (PFE)~\cite{pfe} that proposed to learn a probabilistic embedding while retaining a deterministic pre-trained model for the task of learning face embeddings. However, this was done in a unimodal setting using only images. In this work, we aim to utilize pre-trained vision-language models while providing probabilistic embeddings for both modalities. The probabilistic embeddings derived from our proposed \texttt{ProbVLM} are consistent with cross-modal learning at the core of pretrained vision-language models.

\myparagraph{Uncertainty Estimation.} These techniques have been widely explored for different tasks in computer vision~\cite{kendall2017uncertainties,blundell2015weight,lakshminarayanan2017simple,laves2020well,nix1994estimating,slurp,bayescap,Nazarovs_2022_CVPR,plex,guo2022uncertainty,roy2022uncertainty,yulearning,rangnekar2023usim,upadhyay2021robustness,sudarshan2021towards,upadhyay2021uncertainty}. Uncertainties can be broadly categorized into aleatoric~\cite{kendall2017uncertainties,Gast_2018_CVPR,bae2021estimating,wang2021bayesian,cui2023bayesmil,ayhan2018test,wang2019aleatoric,nix1994estimating,xie2022towards} and epistemic~\cite{graves2011practical,blundell2015weight,lakshminarayanan2017simple,welling2011bayesian,gal2016dropout,swag,franchi2021one,franchi2020tradi} uncertainties.  
Uncertainty estimation has been used for a variety of tasks, such as identifying model failure~\cite{emami1988effect,Besnier_2021_ICCV,besnier2021learning,whitehead2022reliable} and is extensively used in active learning to select the best samples to train the model~\cite{settles2009active,kirsch2019batchbald,raj2022convergence,shapeev2020active,yang2015multi,yang2016active,prabhu2019sampling,munagala2022clactive}. 
While several of these methods focus on training a new Bayesian model from scratch for quantifying the uncertainties in the prediction, some recent works like~\cite{bayescap,slurp,gradood} have proposed methods to estimate the uncertainties for the pre-trained frozen models. However, these works tackle data from a single modality.
This work efficiently estimates the uncertainty for the pre-trained frozen large-scale vision-language model.

\section{Method}
\label{sec:method}

We first describe the problem formulation in Section~\ref{sec:prob_form}. In Section~\ref{sec:probvlm}, we describe our proposed method \texttt{ProbVLM} that estimates the complex probability distributions for the embeddings of the frozen deterministic vision-langue encoders,
quantifying the uncertainties for their predictions.

\subsection{Problem Formulation}
\label{sec:prob_form}
Let $\mathcal{D}=(\mathcal{I},\mathcal{C})$ denote a vision and language dataset, where $\mathcal{I}$ is a set of images and $\mathcal{C}$ a set of captions 
The two sets are connected via ground-truth matches where multiplicity is plausible. For a caption $c \in \mathcal{C}$ (respectively an image $i \in \mathcal{I}$), the set of corresponding images (respectively captions) is given by $\kappa(c)\subseteq \mathcal{I}$ (respectively $\kappa(i) \subseteq \mathcal{C}$). 
Recent advances in cross-modal vision-language models~\cite{clip,slip,flava} often involve learning a shared embedding space, $\mathcal{Z} \subseteq \mathbb{R}^D$ ($D$-dimensional space), for images and texts. This allows quantifying the similarity between cross-modal elements based on their distances in the shared embedding space. The shared embedding space is learned via a set of two encoders: $\mathbf{\Phi}_{\mathcal{V}}(\cdot; \theta_{\mathcal{V}}): \mathcal{I} \rightarrow \mathcal{Z}$ for the images and $\mathbf{\Phi}_{\mathcal{T}}(\cdot; \theta_{\mathcal{T}}): \mathcal{C} \rightarrow \mathcal{Z}$ for the texts, where $\theta_{\mathcal{V}}$ and $\theta_{\mathcal{T}}$ are the parameters for the respective mapping functions.

We consider a real-world scenario where the above set of encoders have already been trained on vast datasets using large models with high computational cost, e.g., CLIP~\cite{clip}, SLIP~\cite{slip}, Flava~\cite{flava} and BLIP~\cite{blip}, are in \textit{frozen state}, i.e., we have $\mathbf{\Phi}_{\mathcal{V}}(\cdot; \theta_{\mathcal{V}}^*)$ and $\mathbf{\Phi}_{\mathcal{T}}(\cdot; \theta_{\mathcal{T}}^*)$, where $\theta_{\mathcal{V}}^*, \theta_{\mathcal{T}}^*$ represents the parameters of the pretrained frozen encoders. 
These encoders are \textit{deterministic} and map an image/text to vectors in the shared space, i.e., given a sample image $\mathbf{x}_\mathcal{V}$ (and similarly sample text $\mathbf{x}_{\mathcal{T}}$), the encoder provides an embedding $\mathbf{z}_{\mathcal{V}} := \mathbf{\Phi}_{\mathcal{V}}(\mathbf{x}_{\mathcal{V}}; \theta_{\mathcal{V}}^*)$ (and similarly,  $\mathbf{z}_{\mathcal{T}} := \mathbf{\Phi}_{\mathcal{T}}(\mathbf{x}_{\mathcal{T}}; \theta_{\mathcal{T}}^*)$). 
However, the point estimates, $\mathbf{z}$, do not capture the ambiguity inherent to these embeddings~\cite{hib,pcme,fang2023uatvr} that are better represented by the probability distribution $P_{\mathbf{z}|\mathbf{x}}$.
Therefore, we propose to estimate $P_{\mathbf{z}|\mathbf{x}}$ for the pretrained model efficiently, using \texttt{ProbVLM}, quantifying the uncertainties of the output without re-training the encoders.

\subsection{Building \textbf{\texttt{ProbVLM}}}
\label{sec:probvlm}
Despite being deterministic, large-scale \textit{frozen} encoders already provide high-quality point estimates. Our proposed method leverages this fact, using the embeddings $\mathbf{z}$ as estimates for the mean of the desired distribution $P_{\mathbf{z}|\mathbf{x}}$, and estimating the remaining parameters.
$P_{\mathbf{z}|\mathbf{x}}$ can be modeled as a parametric distribution
$
P_{\mathbf{z}|\mathbf{x}}
(
    \mathbf{z} | \{ \hat{\mathbf{z}}, \hat{\mathbf{\nu}} ... \hat{\mathbf{\rho}} \}
)
$
where the parameters can be estimated using a deep neural network~\cite{gal2016dropout,kendall2017uncertainties,lakshminarayanan2017simple}. 
Therefore, we introduce \texttt{ProbVLM}, 
\begin{gather}
\mathbf{\Psi}(\cdot; \zeta): = \left( \mathbf{\Psi}_{\mathcal{V}}(\cdot; \zeta_{\mathcal{V}}), \mathbf{\Psi}_{\mathcal{T}}(\cdot; \zeta_{\mathcal{T}}) \right)
\end{gather}
where $\mathbf{\Psi}_{\mathcal{V}}$ and $\mathbf{\Psi}_{\mathcal{T}}$ represents the vision and text encoders parameterized by $\zeta_{\mathcal{V}}$ and $\zeta_{\mathcal{T}}$, respectively. Also, $\zeta := \zeta_{\mathcal{V}} \cup \zeta_{\mathcal{T}}$ represents the overall parameters for \texttt{ProbVLM}.
that learns to estimate the parameters $\{ \hat{\mathbf{z}}, \hat{\mathbf{\nu}} ... \hat{\mathbf{\rho}} \}$ with the help of frozen encoders $\mathbf{\Phi}_{\mathcal{V}}(\cdot; \theta_{\mathcal{V}}^*)$ and 
 $\mathbf{\Phi}_{\mathcal{T}}(\cdot; \theta_{\mathcal{T}}^*)$. 
 The functions $\mathbf{\Psi}_{\mathcal{V}}(\cdot; \zeta_{\mathcal{V}})$ and $\mathbf{\Psi}_{\mathcal{T}}(\cdot; \zeta_{\mathcal{T}})$ operate on image and text embeddings respectively, but during training depend on both modalities, as discussed later. 
 We design the learning scheme for $\mathbf{\Psi}(\cdot; \zeta)$ such that:
 (i)~Estimated parameter $\hat{\mathbf{z}}$ should remain faithful to the original unimodal embedding $\mathbf{z}$
 (\textit{intra-modal alignment}), this makes the uncertainty of the \texttt{ProbVLM} serve as a good proxy for the uncertainty of frozen encoders.
 (ii)~Estimated parameters $\{ \hat{\mathbf{\nu}} ... \hat{\mathbf{\rho}} \}$ should capture the ambiguities and uncertainties present within and across modalities (\textit{cross-modal alignment}). Figure~\ref{fig:arch} depicts \texttt{ProbVLM} in tandem with the frozen VLM. 

 \myparagraph{Intra-modal Alignment.}
 To ensure that the mean of the distribution estimated by $\mathbf{\Psi}(\cdot; \zeta)$ reflects the point estimates provided by the frozen encoders, we set up a probabilistic reconstruction problem for the embeddings within the modalities. That is, for a given sample $\mathbf{x}$ (either from image or text modality), we obtain the embedding from the frozen encoder $\mathbf{z} = \mathbf{\Phi}(\mathbf{x}; \theta)$ (using the appropriate encoder), then the modality-specific component of $\mathbf{\Psi}(\cdot; \zeta)$ learns to reconstruct the $\mathbf{z}$ (let the reconstruction be called $\hat{\mathbf{z}}$). The modality-specific component of $\mathbf{\Psi}(\cdot; \zeta)$ is designed to 
 (i)~relax the i.i.d constraints by assuming independent but \textit{not} identically distributed residuals 
 and 
 (ii)~learn the \textit{heteroscedasticity} for the residuals at the time of reconstruction that may follow the heavy-tailed distributions~\cite{bayescap,srganawate,kumar2017kernel,kumar2017kernel1,cmu_cite}. 
 The modality-specific component is learned by maximizing the likelihood, $\mathcal{L}(\zeta; \{\mathbf{z}_i\}_{i=1}^{N})$ for the embeddings of $N$ samples in the datasets. That is, the modality-specific optimal parameters are given by,
 \begin{gather}
     \zeta^* := \underset{\zeta}{\text{argmax }} \mathcal{L}(\zeta; \{\mathbf{z}_i\}_{i=1}^{N}) = \prod_{i=1}^{N}  \frac{\hat{\mathbf{\beta}}_i e^{- (|\hat{\mathbf{z}}_i - \mathbf{z}_i|/\hat{\mathbf{\alpha}}_i)^{\hat{\mathbf{\beta}}_i}} }{2 \hat{\mathbf{\alpha}}_i \Gamma(1/\hat{\mathbf{\beta}}_i)} 
     \label{eq:ggd}
 \end{gather}
 In the above equation, $ \frac{\hat{\mathbf{\beta}}_i e^{- (|\hat{\mathbf{z}}_i - \mathbf{z}_i|/\hat{\mathbf{\alpha}}_i)^{\hat{\mathbf{\beta}}_i}} }{2 \hat{\mathbf{\alpha}}_i \Gamma(1/\hat{\mathbf{\beta}}_i)}$ represents the \textit{generalized Gaussian distribution} (GGD, represented by $\mathcal{G}$) that is capable of modeling heavy-tailed distributions (note the Gaussian and Laplace are special cases of $\mathcal{G}$ with $\alpha=1, \beta=2$ and $\alpha=1, \beta=1$, respectively).
 The variables $\hat{\mathbf{z}}_i, \hat{\mathbf{\alpha}}_i, \hat{\mathbf{\beta}}_i$ are the predicted mean, scale, and shape parameters of $\mathcal{G}$ from our modality-specific components for the given input $\mathbf{z}_i$.
We obtain modality-specific optimal parameters by minimizing negative log-likelihood (equivalent to Equation~\ref{eq:ggd}). Given $\mathbf{z}$ and predicted $\hat{\mathbf{z}}, \hat{\mathbf{\alpha}}, \hat{\mathbf{\beta}}$, loss is given by,
 \begin{gather}
     L_{\text{rec}}(\zeta) := \left( \frac{|\hat{\mathbf{z}} - \mathbf{z}|}{\hat{\mathbf{\alpha}}} \right)^{\hat{\mathbf{\beta}}} - \log \frac{\hat{\mathbf{\beta}}}{\hat{\mathbf{\alpha}}} + \log \Gamma(\frac{1}{\hat{\mathbf{\beta}}})
     \label{eq:intra_nll}
 \end{gather}
 Therefore, the vision-specific component of \texttt{ProbVLM}, $\mathbf{\Psi}(\cdot; \zeta_{\mathcal{V}})$),
 is trained by minimizing the Eqation~\ref{eq:intra_nll} using image embeddings, we denote this loss as $L_{\text{rec}}^{\mathcal{V}}(\zeta_V)$. Similarly the text-specific component, $\mathbf{\Psi}(\cdot; \zeta_{\mathcal{T}})$, is trained by minimizing $L_{\text{rec}}^{\mathcal{T}}(\zeta_T)$.
 As discussed next, we also enforce cross-modal alignment so that the predicted distribution of \texttt{ProbVLM} captures the uncertainties across modalities from one-to-many correspondences for an embedding.

 \myparagraph{Cross-modal Alignment.}
 While the intra-modal alignment seeks to match the means of the output distribution from \texttt{ProbVLM} to the embeddings derived from frozen vision-language encoders, 
 we also enforce the image and text embedding output distribution (from \texttt{ProbVLM}) belonging to similar concepts to remain close to each other. 
 That is, given an image and text embedding pair $(\mathbf{z}_\mathcal{V}, \mathbf{z}_\mathcal{T})$ (from frozen model) representing similar concepts, the output distributions from $\mathbf{\Psi}(\cdot; \zeta)$, 
 $\mathcal{G}(\mathbf{z}; \hat{\mathbf{z}}_{\mathcal{V}}, \hat{\mathbf{\alpha}}_{\mathcal{V}}, \hat{\mathbf{\beta}}_{\mathcal{V}})$ and $\mathcal{G}(\mathbf{z}; \hat{\mathbf{z}}_{\mathcal{T}}, \hat{\mathbf{\alpha}}_{\mathcal{T}}, \hat{\mathbf{\beta}}_{\mathcal{T}})$ (later referred to as $\mathcal{G}_{\mathcal{V}}(\mathbf{z})$) and $\mathcal{G}_{\mathcal{T}}(\mathbf{z})$) should match. 
 This can be measured directly from the likelihood as, $p(\mathbf{z}_v = \mathbf{z}_u)$, where 
$\mathbf{z}_v \sim \mathcal{G}_{\mathcal{V}}(\mathbf{z})$ and
$\mathbf{z}_u \sim \mathcal{G}_{\mathcal{T}}(\mathbf{z})$ as in~\cite{pfe} 
, i.e.,
\begin{gather}
    \scalebox{1.0}{$
    p(\mathbf{z}_v = \mathbf{z}_u)
    :=$} 
    \iint  
    \scalebox{0.99}{$
    \mathcal{G}_{\mathcal{V}}(\mathbf{z}_v) \mathcal{G}_{\mathcal{T}}(\mathbf{z}_u) \delta(\mathbf{z}_v-\mathbf{z}_u) d\mathbf{z}_v d\mathbf{z}_u
    $}
\end{gather}
where $\delta(\cdot)$ refers to the \textit{Dirac-delta distribution}.
The above integral can be simplified further by defining $\Delta \mathbf{z} = \mathbf{z}_v - \mathbf{z}_u$ and seeking $p(\Delta \mathbf{z}) = 0$. As both $\mathbf{z}_v$ and $\mathbf{z}_u$ are GGD random variables, $\Delta \mathbf{z}$ follows the distribution based on the \textit{Bivariate Fox H-function}~\cite{BFHF,hf1,hf2} given by,
\begin{gather}
    \Delta z \sim \scalebox{0.9}{$ \frac{1}{2 \Gamma(1/\hat{\mathbf{\beta}}_{\mathcal{V}}), \Gamma(1/\hat{\mathbf{\beta}}_{\mathcal{T}})} $} \times
     \nonumber \\
    \int
    \scalebox{0.8}{$
    \mathcal{H}^{1,1}_{1,2} \left[ A t^2 \big| 
 \begin{matrix}
 (1-\frac{1}{\hat{\mathbf{z}}_{\mathcal{V}}}, \frac{1}{\hat{\mathbf{z}}_{\mathcal{T}}}) \\
 (0,1) (\frac{1}{2}, 1)
 \end{matrix}
 \right]
    \mathcal{H}^{1,1}_{1,2} \left[ B t^2 \big| 
 \begin{matrix}
 (1-\frac{1}{\hat{\mathbf{z}}_{\mathcal{T}}}, \frac{1}{\hat{\mathbf{z}}_{\mathcal{T}}}) \\
 (0,1) (\frac{1}{2}, 1)
 \end{matrix}
 \right] \cos{t(\mu - z)}
    $}
    dt
    \label{eq:bfhf}
\end{gather}
Where $A=\frac{\hat{\mathbf{\alpha}}_{\mathcal{V}}^2 \Gamma(1/\hat{\mathbf{\beta}}_{\mathcal{V}})}{4 \Gamma(3/\hat{\mathbf{\beta}}_{\mathcal{V}})}$, $B=\frac{\hat{\mathbf{\alpha}}_{\mathcal{T}}^2 \Gamma(1/\hat{\mathbf{\beta}}_{\mathcal{T}})}{4 \Gamma(3/\hat{\mathbf{\beta}}_{\mathcal{T}})}$, $\mu=\hat{\mathbf{z}}_v - \hat{\mathbf{z}}_u$, and $\mathcal{H}$ is the \textit{Fox H function}~\cite{BFHF,hf1,hf2}.
Equation~\ref{eq:bfhf} does not provide a scalable objective function suitable for training deep neural networks. 
Hence, we propose an approximation that is easily scalable for deep-learning models given by,
\begin{gather}
    p(\mathbf{z}_v = \mathbf{z}_u) =
    \iint  
    \mathcal{G}_{\mathcal{V}}(\mathbf{z}_v)
    \mathcal{G}_{\mathcal{T}}(\mathbf{z}_u)
    \delta(\mathbf{z}_v-\mathbf{z}_u) d\mathbf{z}_v d\mathbf{z}_u \nonumber \\
    \approx \int 
    \frac{1}{2}
    \left(
    \mathcal{G}_{\mathcal{V}}(\mathbf{z}) \delta(\mathbf{z} - \mathbf{z}_{\mathcal{T}}) + 
    \mathcal{G}_{\mathcal{T}}(\mathbf{z}) \delta(\mathbf{z} - \mathbf{z}_{\mathcal{V}})
    \right)
    d\mathbf{z}
    \label{eq:cm_like}
\end{gather}
The appendix shows details of the above equation.
The first term in the integral, $\int \mathcal{G}_{\mathcal{V}}(\mathbf{z}) \delta(\mathbf{z} - \mathbf{z}_{\mathcal{T}}) d\mathbf{z}$, is the likelihood of the text embedding $\mathbf{z}_{\mathcal{T}}$ under the predicted distribution, $\mathcal{G}_{\mathcal{V}}(\mathbf{z})$, for the visual embedding.
Similarly, the second term is the likelihood of the visual embedding $\mathbf{z}_{\mathcal{V}}$ under the predicted distribution, $\mathcal{G}_{\mathcal{T}}(\mathbf{z})$, for the text embedding.
Negative log of Equation~\ref{eq:cm_like} leads to a scalable objective function that can be used to learn the optimal parameters for vision and text components of \texttt{ProbVLM} ($\mathbf{\Psi}_{\mathcal{V}}(\cdot; \zeta_{\mathcal{V}})$ and $\mathbf{\Psi}_{\mathcal{T}}(\cdot; \zeta_{\mathcal{T}})$),
\begin{gather}
    L_{\text{cross}}(\zeta_{\mathcal{V}}, \zeta_{\mathcal{T}}) :=  %
    \scalebox{0.9}{$
    \underbrace{
    \left( \frac{|\hat{\mathbf{z}}_{\mathcal{V}} - \mathbf{z}_{\mathcal{T}}|}{\hat{\mathbf{\alpha}}_{\mathcal{V}}} \right)^{\hat{\mathbf{\beta}}_{\mathcal{V}}} - \log \frac{\hat{\mathbf{\beta}}_{\mathcal{V}}}{\hat{\mathbf{\alpha}}_{\mathcal{V}}} + \log \Gamma(\frac{1}{\hat{\mathbf{\beta}}_{\mathcal{V}}})
    }_{\text{Cross-modal: vision} \rightarrow \text{text}}
    $}
    + \nonumber \\
    \scalebox{0.9}{$
    \underbrace{
    \left( \frac{|\hat{\mathbf{z}}_{\mathcal{T}} - \mathbf{z}_{\mathcal{V}}|}{\hat{\mathbf{\alpha}}_{\mathcal{T}}} \right)^{\hat{\mathbf{\beta}}_{\mathcal{T}}} - \log \frac{\hat{\mathbf{\beta}}_{\mathcal{T}}}{\hat{\mathbf{\alpha}}_{\mathcal{T}}} + \log \Gamma(\frac{1}{\hat{\mathbf{\beta}}_{\mathcal{T}}})
    }_{\text{Cross-modal: text} \rightarrow \text{vision}}
    $}
\end{gather}
The overall objective used for \texttt{ProbVLM} is designed to be,
\begin{gather}
    \scalebox{0.88}{$
    L_{\text{ProbVLM}}(\zeta_{\mathcal{V}}, \zeta_{\mathcal{T}}) = L_{\text{rec}}^{\mathcal{V}}(\zeta_{\mathcal{V}}) + L_{\text{rec}}^{\mathcal{T}}(\zeta_{\mathcal{T}}) + \lambda_{\text{cross}}  L_{\text{cross}}(\zeta_{\mathcal{V}}, \zeta_{\mathcal{T}})
    $}
\end{gather}
where $\lambda_{cross}$ is a hyperparameter controlling the relative contribution of inter-intra modality terms. 

\begin{figure*}[t]
    \centering
    \includegraphics[width=0.5\textwidth]{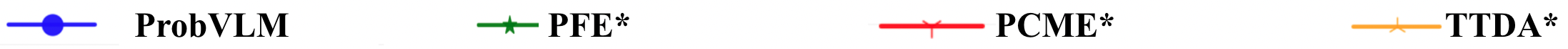}
  \includegraphics[width=0.93\textwidth]{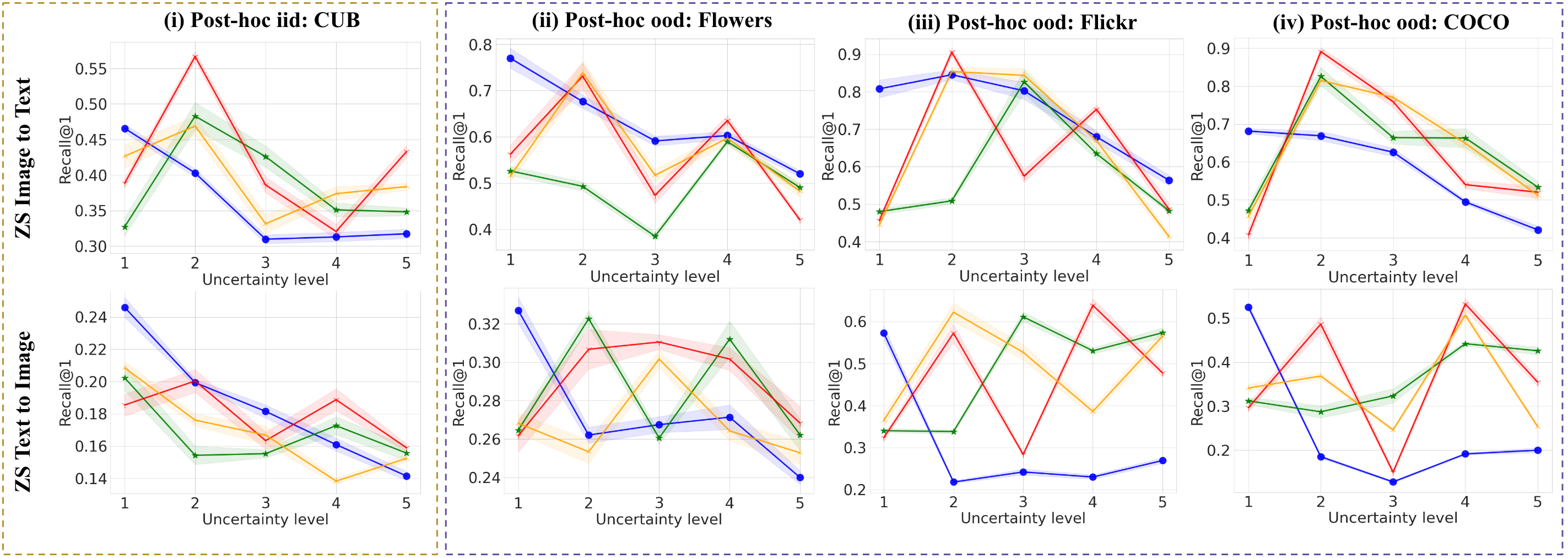}
  \includegraphics[width=0.93\textwidth]{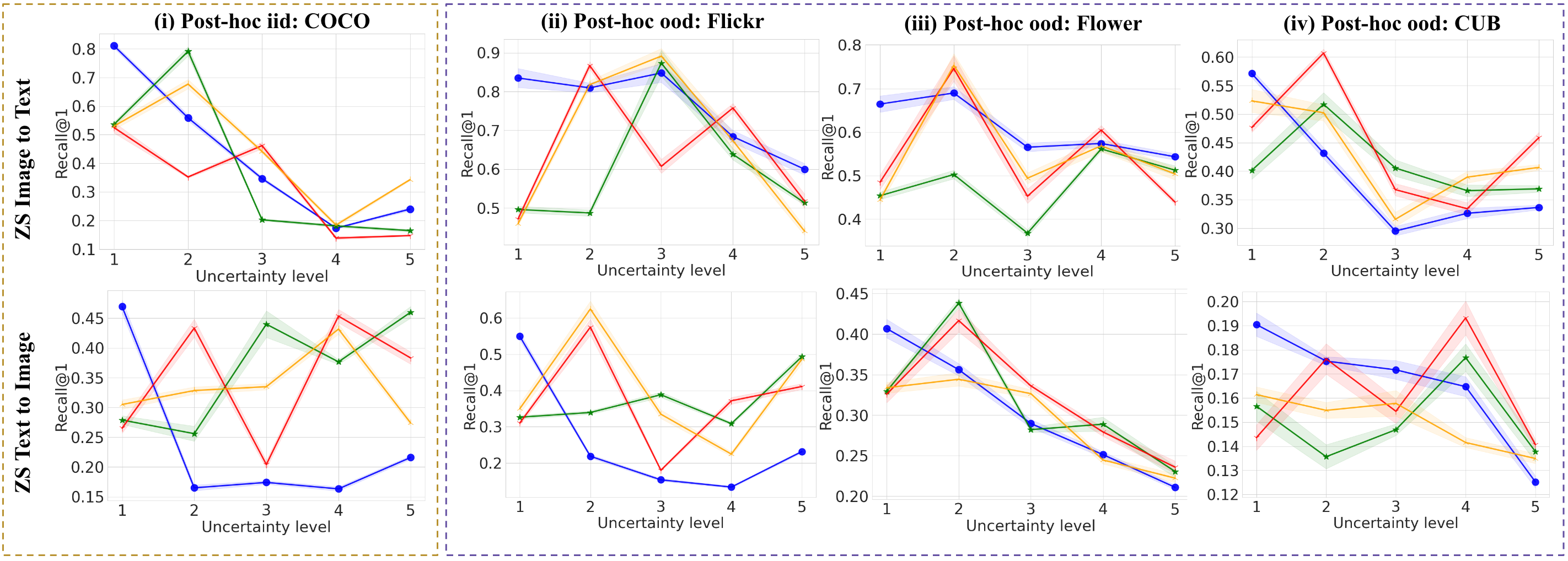}
    \vspace{-10pt}
    \caption{Measuring the calibration with various post-hoc method for Image-to-Text and Text-to-Image retrieval when trained on (top) CUB and (bottom) COCO, and evaluated on CUB, COCO, Flickr, FLO.}
    \label{fig:calib}
    \vspace{-15pt}
\end{figure*}

\myparagraph{Uncertainty Quantification.} 
Given embedding $\mathbf{z}$ from a frozen encoder, predicted distributions from the trained \texttt{ProbVLM} (output from the appropriate component) allows aleatoric uncertainty estimation as $\hat{\mathbf{\sigma}}_{\text{aleatoric}}^2 = \frac{\hat{\mathbf{\alpha}}^2 \Gamma(3/\hat{\mathbf{\beta}})}{\Gamma(1/\hat{\mathbf{\beta}})}$.
Moreover, we design both $\mathbf{\Psi}_{\mathcal{V}}$ and $\mathbf{\Psi}_{\mathcal{T}}$ to be simple 3-layer MLPs with dropout layers (with dropout probability set to 0.1 during training). Activating dropouts during inference, with multiple forward passes (say $M$), allows estimating the epistemic uncertainty, 
$\hat{\mathbf{\sigma}}^2_{\text{epistemic}} = \frac{1}{M}\sum_{m=1}^M(\hat{\mathbf{z}}_m - \frac{1}{M}\sum_{j=1}^M\hat{\mathbf{z}}_j)^2$. We estimate total uncertainty as,
\begin{gather}
    \hat{\mathbf{\sigma}}_{\text{total}}^2 = \hat{\mathbf{\sigma}}^2_{\text{epistemic}} + \hat{\mathbf{\sigma}}_{\text{aleatoric}}^2
\end{gather}

\subsection{Latent Diffusion for Probabilistic Embeddings}
\label{sec:st}
For a given text embedding $\mathbf{z}_{\mathcal{T}}$, the distribution estimated via \texttt{ProbVLM}, $\mathcal{G}(\mathbf{z}; \hat{\mathbf{z}}_{\mathcal{T}}, \hat{\mathbf{\alpha}}_{\mathcal{T}}, \hat{\mathbf{\beta}}_{\mathcal{T}})$ can be visualized by drawing samples from the predicted distribution of vectors (say, $\{\hat{\mathbf{z}}_{\mathcal{T},i}\}_{i=1}^{Q}$) and passing them through a latent diffusion model, e.g., \textit{Stable Diffusion} (say, $\mathbf{\Omega}(\cdot; \theta_{\Omega}^*)$) using CLIP text encoder, to synthesize the set of samples (say, $J$) from the corresponding distribution of images, i.e., 
\begin{gather}
    J := \{\mathbf{\Omega}(\hat{\mathbf{z}}_i; \theta_{\Omega})\}_{i=1}^{Q}
\end{gather}
Section~\ref{sec:diffusion} uses this to visualize the predicted distributions.

\section{Experiments and Results}
\label{sec:exp}
We start by highlighting our tasks, datasets, and evaluation metrics. We also compare our model to various state-of-the-art methods quantitatively and qualitatively in Section~\ref{sec:res}. In Section~\ref{sec:abl}, we provide an ablation analysis, and Section~\ref{sec:app} demonstrates some real-world applications of \texttt{ProbVLM} for model selection and active learning.

{
\setlength{\tabcolsep}{3pt}
\renewcommand{\arraystretch}{1.4}
\begin{figure}[t]
\resizebox{\linewidth}{!}{
\centering
\begin{tabular}{l l l cccc cccc}
    &   &   & \multicolumn{4}{c}{\textbf{i2t}} & \multicolumn{4}{c}{\textbf{t2i}} \\\cmidrule(lr){4-7} \cmidrule(lr){8-11}
    \textbf{VL}  & \textbf{M} & \textbf{Metrics} & \textbf{COCO} & \textbf{Flickr} & \textbf{FLO} & \textbf{CUB} & \multicolumn{1}{l}{\textbf{COCO}} & \multicolumn{1}{l}{\textbf{Flickr}} & \multicolumn{1}{l}{\textbf{FLO}} & \textbf{CUB} \\ 
    \midrule
    \multirow{12}{*}{\rotatebox[origin=c]{90}{CLIP}} & \multirow{3}{*}{\rotatebox[origin=c]{90}{ProbVLM}} & S $\downarrow$ & \textbf{-0.99}  & \textbf{-0.70}  & \textbf{-0.90} & \textbf{-0.60} & \textbf{-0.30} & \textbf{-0.70} & \textbf{-0.99} & \textbf{-0.89}   \\  
    &  & R$^2 \uparrow$ & \textbf{0.93} & \textbf{0.71} & \textbf{0.62} & \textbf{0.67} & \textbf{0.35} &  \textbf{0.50} & \textbf{0.99} &   \textbf{0.70}  \\  
    &  & -SR$^2 \uparrow$ &  \textbf{0.93} & \textbf{0.49}  & \textbf{0.56} & \textbf{0.40} & \textbf{0.10} & \textbf{0.35} & \textbf{0.99} & \textbf{0.63}  \\ 
    \cmidrule(l){2-11} 
    & \multirow{3}{*}{\rotatebox[origin=c]{90}{PFE*\cite{pfe}}} & S $\downarrow$ & -0.79  & -0.19  & 0.60  & \textbf{-0.60} & 0.79 & 0.30 & -0.89 & -0.10    \\  
    && R$^2 \uparrow$   &   0.59 & 0.01  &  0.30 & 0.28&  0.74 & 0.44 & 0.52  &  0.00   \\  
    && -SR$^2 \uparrow$ &  0.47 & 0.00  & -0.18  &  0.17& -0.59  &  -0.13  &  0.47&  -0.00   \\ 
    \cmidrule(l){2-11} 
    & \multirow{3}{*}{\rotatebox[origin=c]{90}{PCME*\cite{pcme}}}   & S $\downarrow$& -0.89 & -0.30  & -0.30  & \textbf{-0.60} & 0.30  &    0.09    &      -0.70   & 0.30    \\  
    && R$^2 \uparrow$   & 0.75  & 0.07  & 0.07  & 0.20  &  0.16  &  0.01   &   0.57  &  0.01   \\  
    && -SR$^2 \uparrow$ &  0.68 & 0.02  & 0.02  &  0.12     & -0.05 &  -0.00     &  0.40 & -0.00    \\ 
    \cmidrule(l){2-11} 
    & \multirow{3}{*}{\rotatebox[origin=c]{90}{TTDA\cite{ayhan2018test}}}    & S $\downarrow$& -0.79  & -0.30  & 0.00  & \textbf{-0.60}     & -0.10  & -0.19   & -0.89 &  -0.50   \\  
    && R$^2 \uparrow$   & 0.69  & 0.09  & 0.00  & 0.41      & 0.26  &  0.071 & 0.80    & 0.15    \\  
    && -SR$^2 \uparrow$ &   0.55 & 0.03  & 0.00  & 0.24 & 0.00  &  0.01     &  0.73    & 0.07    \\ 
    \midrule
    \multirow{12}{*}{\rotatebox[origin=c]{90}{BLIP}} & \multirow{3}{*}{\rotatebox[origin=c]{90}{ProbVLM}} & S $\downarrow$& \textbf{-0.87}  & \textbf{-0.79}  & \textbf{-0.74}  &    \textbf{-0.66}   &    \textbf{-0.43}    &     \textbf{-0.38}      &    \textbf{-0.31} & \textbf{-0.22}    \\  
    && R$^2 \uparrow$   &  \textbf{0.92} & \textbf{0.83}  & \textbf{0.68}  &  \textbf{ 0.61}    & \textbf{0.52}&      \textbf{0.48}     &     \textbf{0.45}& \textbf{0.38}    \\  
    && -SR$^2 \uparrow$ & \textbf{0.80}  & \textbf{0.66}  & \textbf{0.50}  &    \textbf{0.40}   & \textbf{0.22}&      \textbf{0.18}   &\textbf{0.14}     &   \textbf{0.08}   \\ 
    \cmidrule(l){2-11} 
    & \multirow{3}{*}{\rotatebox[origin=c]{90}{PFE*\cite{pfe}}}    & S $\downarrow$& -0.82 & -0.74	& -0.63	& -0.63	&	-0.39 &	-0.32 &	-0.28 &	-0.18 \\  
    && R$^2 \uparrow$   & 0.72 &	0.76 &	0.62 &	0.44	&	0.48 &	0.38 &	0.39 &	0.37   \\  
    && -SR$^2 \uparrow$ & 0.58 &	0.57 &	0.39 &	0.27	&	0.19 &	0.12 &	0.11 &	0.07  \\ 
    \cmidrule(l){2-11} 
    & \multirow{3}{*}{\rotatebox[origin=c]{90}{PCME*\cite{pcme}}}   & S $\downarrow$& -0.76 & -0.53 &	-0.60 &	-0.44 &		-0.28 &	-0.26 &	-0.28 &	-0.21\\  
    && R$^2 \uparrow$   &   0.81 &	0.56 &	0.60 &	0.53 &		0.50 &	0.34 &	0.44 &	0.36     \\  
    && -SR$^2 \uparrow$ &  0.62 &	0.29 &	0.36 &	0.23 &		0.14 &	0.09 &	0.12 &	\textbf{0.08}     \\ 
    \cmidrule(l){2-11} 
    & \multirow{3}{*}{\rotatebox[origin=c]{90}{TTDA\cite{ayhan2018test}}}    & S $\downarrow$& -0.44 &	-0.33 &	-0.74 &	-0.60 &		-0.19 &	-0.26 &	-0.21 &	-0.21 \\  
    && R$^2 \uparrow$  & 0.66 &	0.56 &	0.42 &	0.55 &		0.49 &	0.23 &	0.35 &	0.36 \\  
    && -SR$^2 \uparrow$ & 0.29 &	0.18 &	0.31 &	0.33 &		0.10 &	0.06 &	0.07 &	\textbf{0.08}\\ 
    \bottomrule
\end{tabular}%
}
\captionof{table}{Metrics to evaluate the calibration of the uncertainty estimates for both CLIP~\cite{clip} and BLIP~\cite{blip} Vision-Language models for all considered methods, trained on COCO and evaluated on COCO, Flickr, CUB, and FLO.}
\label{tab:calibration}
\vspace{-8pt}
\end{figure}
}

\myparagraph{Datasets, Metrics, and Baselines.} We use the MS-COCO~\cite{mscoco}, Flickr-30k~\cite{flickr30k}, and the CUB~\cite{cub} as they are widely used for cross-modal retrieval~\cite{pcme,vse,pvse}. Furthermore, we adapt the Oxford-Flowers 102 (FLO) dataset~\cite{flo} similar to \cite{pcme} as an additional benchmark for cross-modal retrieval in a fine-grained setting. We evaluate the performance of both Image-to-Text retrieval and Text-to-Image Retrieval using the Recall@k (R@k) metric. To evaluate the calibration of the uncertainty estimates, we define uncertainty levels~\cite{pcme} and use the Spearman rank correlation (denoted by $S$) between the uncertainty level and the Recall@k for retrieval. For an ideal model, performance would decrease monotonically with increasing uncertainty levels, leading to a score of -1.
We also compute the $R^2$ for the regression fit between the uncertainty levels and R@1 performances to measure if the drop in performance follows a linear trend. Finally, we also measure the product of these two scores (as a unified metric), i.e., $-SR^2$, which should be 1.0 for an ideal model. 
Since there is \textit{no prior work} to estimate probabilistic embeddings from a deterministic model in a cross-modal setting, we adapt a few existing ideas for the task. 
The first baseline is adapted from PFE~\cite{pfe}, where we learn the covariances for the heteroscedastic Gaussian distribution while keeping the mean fixed to the embeddings derived from the frozen encoders in each modality.
The second is to use the soft-contrastive objective of PCME\cite{pcme} to train a probabilistic adapter in a post-hoc fashion. 
Finally, we also have a baseline that performs perform Test-Time Data Augmentation (TTDA) on the inputs~\cite{ayhan2018test,wang2019aleatoric}. This is done by perturbing the images and masking out words in the text. While TTDA does not require additional training, we train our \texttt{ProbVLM} and other baselines on datasets like COCO, Flickr, CUB, and FLO.

\myparagraph{Implementation Details.} Our \texttt{ProbVLM} consists of a Multi-Layer Perceptron (MLP) for both the image and text encoder, each consisting of an input layer going from embedding dimension to 256, a hidden layer of size 256, and an output layer going from 256 to embedding dimensions. This is trained for 100 epochs with a learning rate of $1e^{-4}$. More details are available in the supplementary.

\subsection{Calibrated Uncertainty via \textbf{\texttt{ProbVLM}}}
\label{sec:res}
We investigate the calibration of the uncertainty derived from \texttt{ProbVLM} for the cross-modal retrieval task. All models trained on CUB and COCO were evaluated on all four datasets. Calibration plots are illustrated in Figure~\ref{fig:calib}. We observe that the R@1 performance consistently drops for \texttt{ProbVLM} as we increase the uncertainty levels, whereas the baselines rarely see a monotonic drop in performance. We quantify this performance in Table~\ref{tab:calibration}. The highest score of 0.93 for $-SR^2$ (i2t) on the COCO dataset indicates a decreasing performance trend with increasing uncertainty. Notably, the uncertainty estimates indicate the average performance in different bins even when \texttt{ProbVLM} is evaluated on datasets that are different from the train set. In some cases, we see that \texttt{ProbVLM} even achieves a nearly perfect score ($-SR^2$ of 0.99, with CLIP VLM on FLO, after training on COCO for Image-to-Text Retrieval). On the contrary, we find that the baselines often achieve poor scores. It is important to note that all these models use the same underlying embeddings and achieve the same performance on the retrieval task. Of all the considered methods, \texttt{ProbVLM} provides the most calibrated uncertainty estimates. 
We see similar trends for \texttt{ProbVLM} with BLIP~\cite{blip}, where \texttt{ProbVLM} achieves a $-SR^2$ of 0.80, when trained on COCO and evaluated on COCO, compared to other methods like PFE$^*$ (0.58), PCME$^*$ (0.62), and TTDA (0.29).

\begin{figure}
    \centering
    \includegraphics[width=0.46\textwidth]{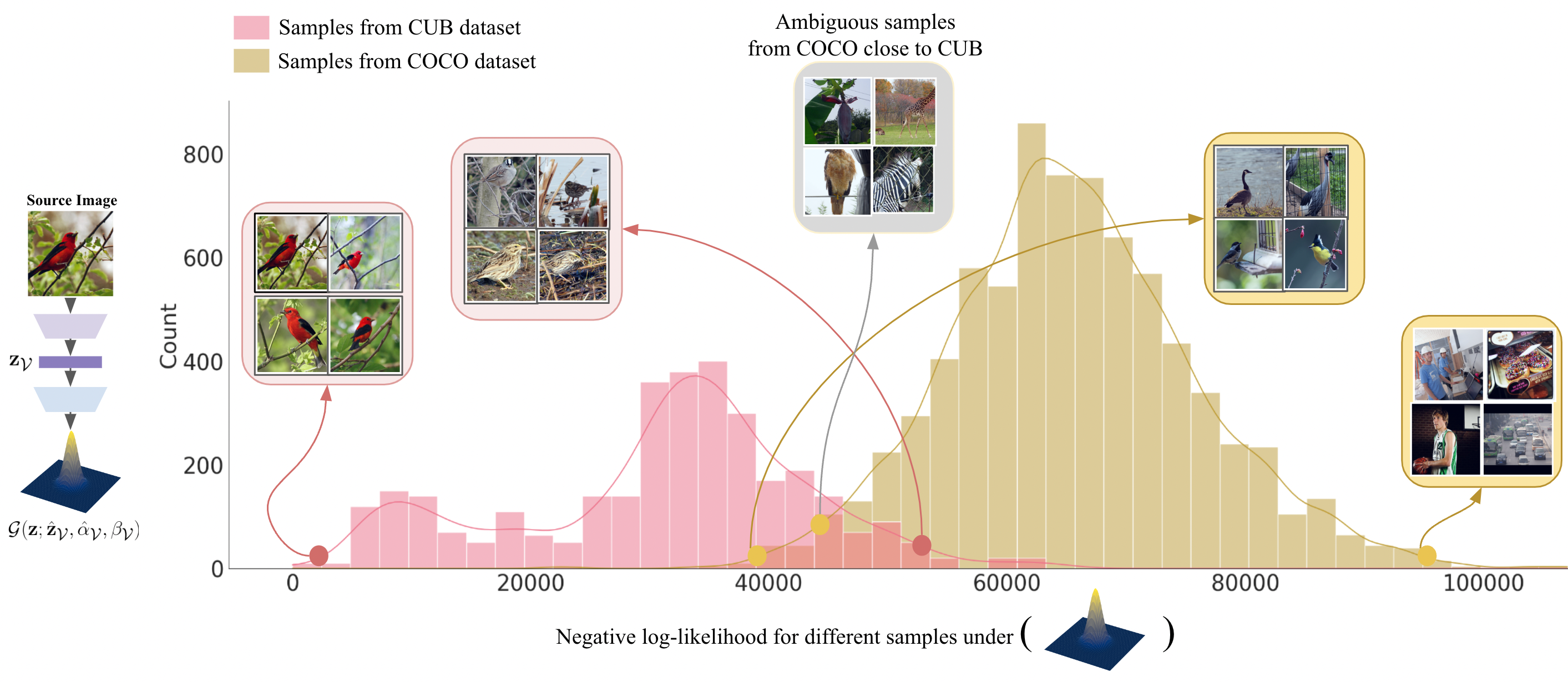}
    \includegraphics[width=0.46\textwidth]{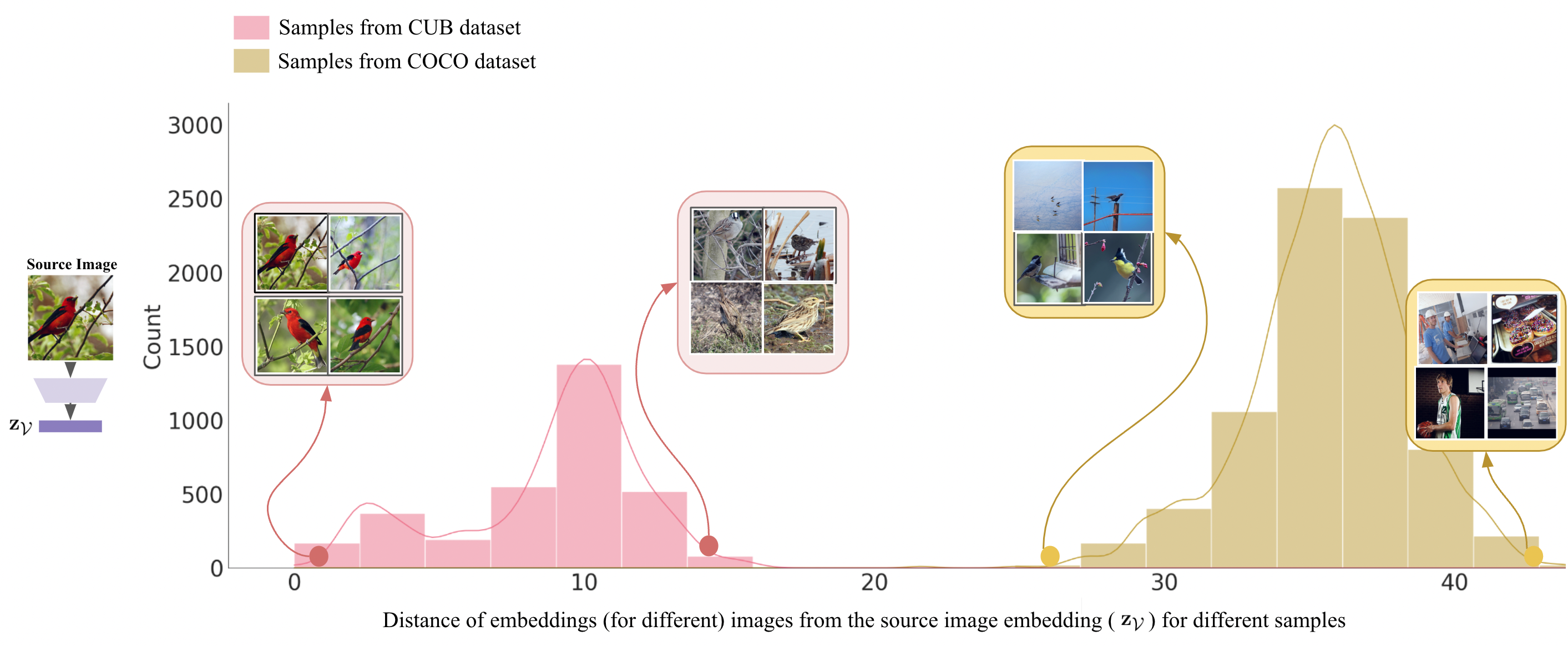}
    \caption{
    Visualizing the uncertainties of the vision encoder captured by \texttt{ProbVLM}. Fixing an image from CUB, we obtain the predicted embedding distribution and compute the likelihood of all other samples in CUB and COCO. We observe that the images in COCO are similar/ambiguous to CUB overlap (Top).
    However, deterministic embeddings lead to a separation between the two datasets (Bottom).%
    }
    \label{fig:nll}
\end{figure}
Figure~\ref{fig:nll}-(Top) visualizes the ambiguities captured by \texttt{ProbVLM} on the visual embeddings. We take a bird image (source) from the CUB dataset and obtain the probability distribution for the visual embedding of that sample; we then compute the likelihood of the visual embeddings (i.e., point estimates derived from CLIP) for the other samples of CUB and COCO datasets, under the source distribution. We notice that within the CUB dataset, the bird images similar to the source image have a higher likelihood compared to other bird images. Also, the images from the COCO dataset tend to have a lower likelihood. However, some images from the COCO dataset have a likelihood similar to the samples from CUB. We visualize these samples and discover them to be bird images.
Moreover, the overlapping region between CUB and COCO has samples from the COCO dataset that are ambiguous and related to bird images as they have similar backgrounds, etc.
On the contrary, when a similar analysis is performed using the CLIP (by measuring the distance between the embeddings instead of likelihood, Figure~\ref{fig:nll}-(Bottom)), we notice that the two datasets are well separated and ambiguities are not captured.

\subsection{Ablations}
\label{sec:abl}
\begin{figure}[t]
    \centering
    \includegraphics[width=0.48\textwidth]{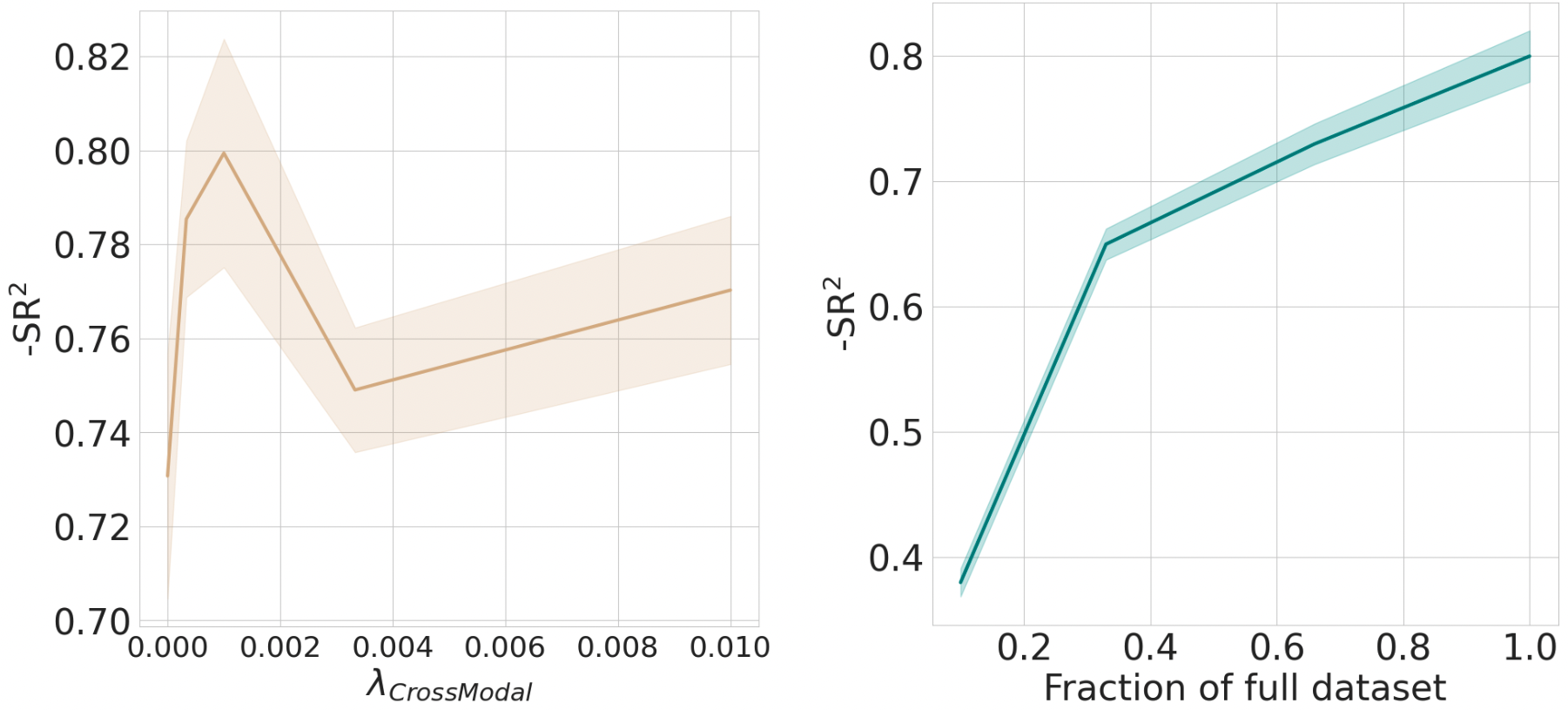}
    \caption{Plot indicating (left)~necessity of the cross-modal alignment and (right)~data required to train \texttt{ProbVLM}.}
    \label{fig:lambda-ablation}
\end{figure}

\begin{figure}[t]
\centering
\includegraphics[width=0.48\textwidth]{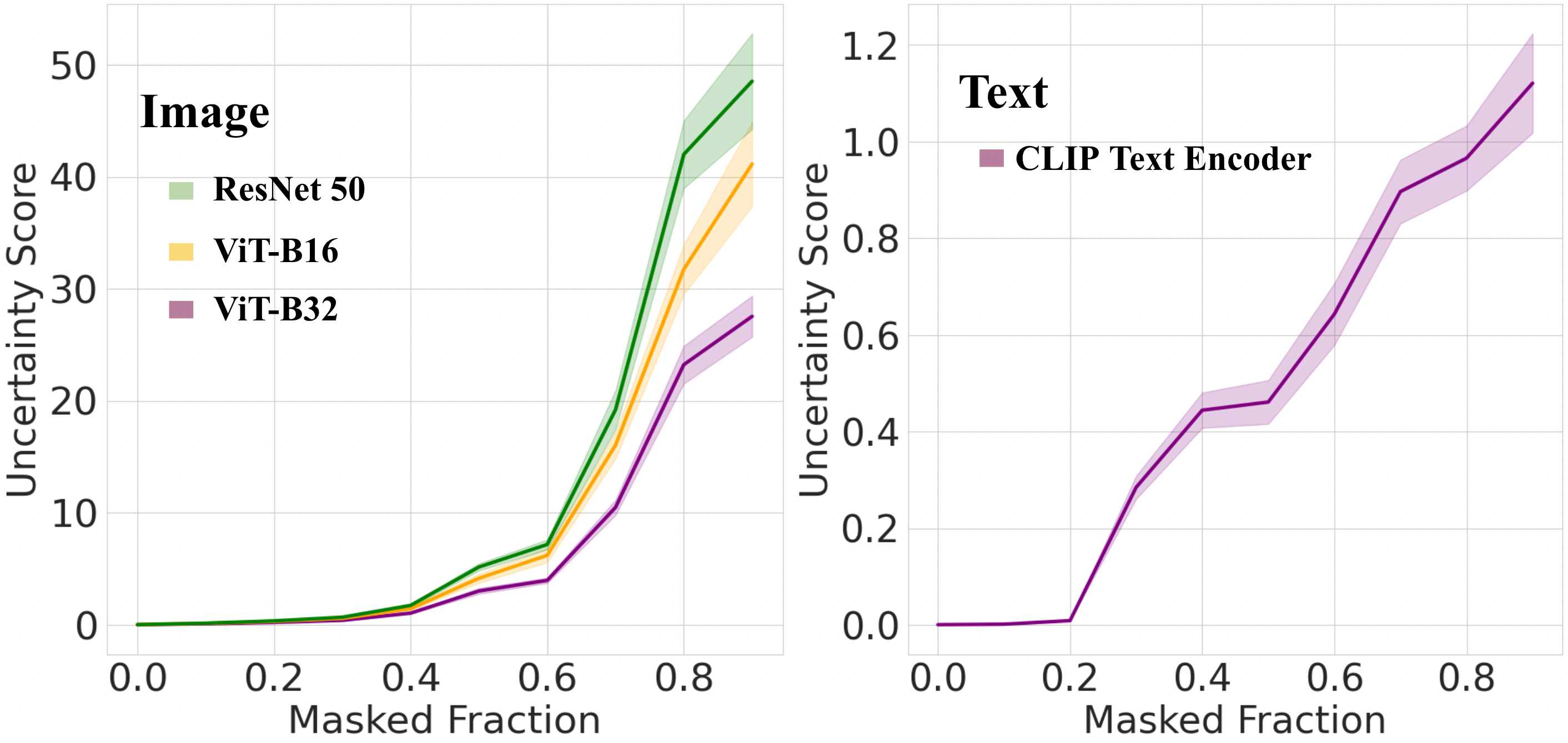}
\caption{Uncertainty increases with increased masking of the input images (Left) and texts (Right). Results with three vision encoders and one language encoder from CLIP.}
\label{fig:masking-effect}
\vspace{-10pt}
\end{figure}

We ablate different components of our proposed \texttt{ProbVLM}, to provide a deeper understanding of its workings. 
First, we perform a sensitivity analysis on the cross-modal reconstruction objective, as shown in Figure~\ref{fig:lambda-ablation}-(Left), for \texttt{ProbVLM} on BLIP using the COCO dataset. We need a non-zero coefficient of the cross-modal loss to ensure that \texttt{ProbVLM} learns meaningful uncertainties that capture the ambiguities across modalities and are well-correlated with its performance on the downstream retrieval task. 
Similarly, having a large co-efficient for the cross-modal loss could hinder learning a faithful identity reconstruction, thereby hampering the performance of the downstream evaluation. 
Next, we investigate the amount of data that is required to train \texttt{ProbVLM} in Figure~\ref{fig:lambda-ablation}-(Right). We get satisfactory calibration of the uncertainty estimates while using only 50\% of the dataset (shown for \texttt{ProbVLM} on BLIP using COCO), indicating that \texttt{ProbVLM} is highly data-efficient. 

Further, we investigate the uncertainties by masking out increasing portions of the input image/text in Figure~\ref{fig:masking-effect}.
We use three different CLIP backbones for the images, ViT-B/32, ViT-B/16, ResNet50, and GPT-based language encoder from CLIP~\cite{clip,gpt2}.
The mean uncertainty steadily increases as we mask increasing amounts of input.

\subsection{Applications}
\label{sec:app}
We study the utility of the uncertainty estimates derived from \texttt{ProbVLM} on two critical applications not well reviewed for VLMs: active learning and model selection.

\begin{figure}[t]
\centering
\includegraphics[width=0.48\textwidth]{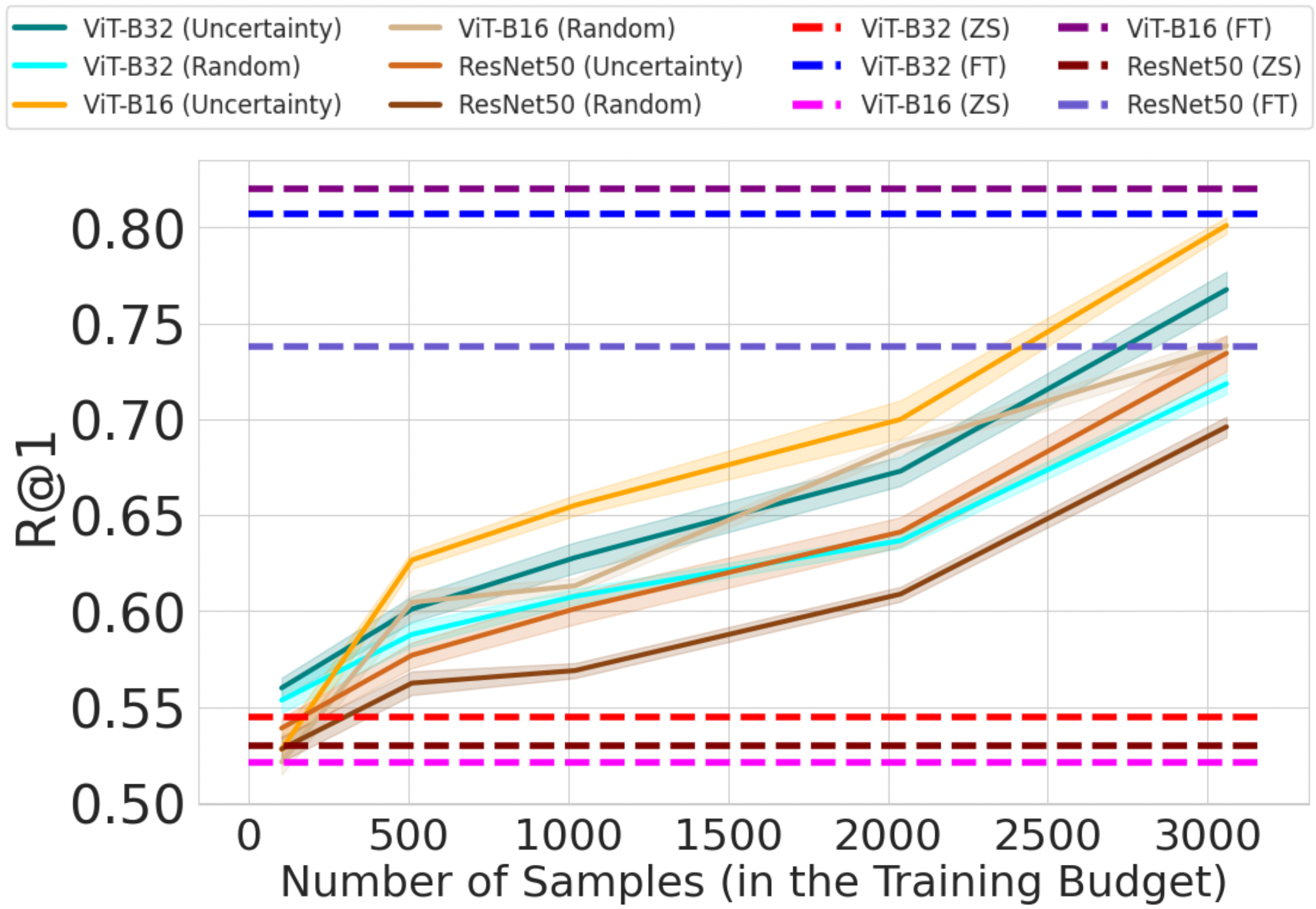}
\caption{Results for active learning, with different vision encoders and varying training budgets. For a given encoder, uncertainty-based sampling outperforms random sampling.}
\label{fig:active}
\vspace{-10pt}
\end{figure}

\begin{figure*}[t]
     \centering
     \includegraphics[width=0.49\textwidth]{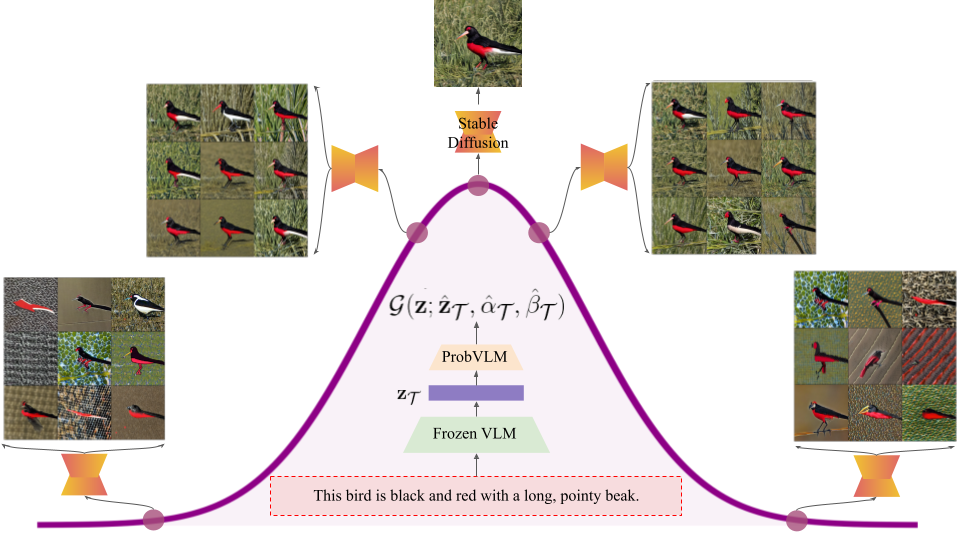}
     \includegraphics[width=0.49\textwidth]{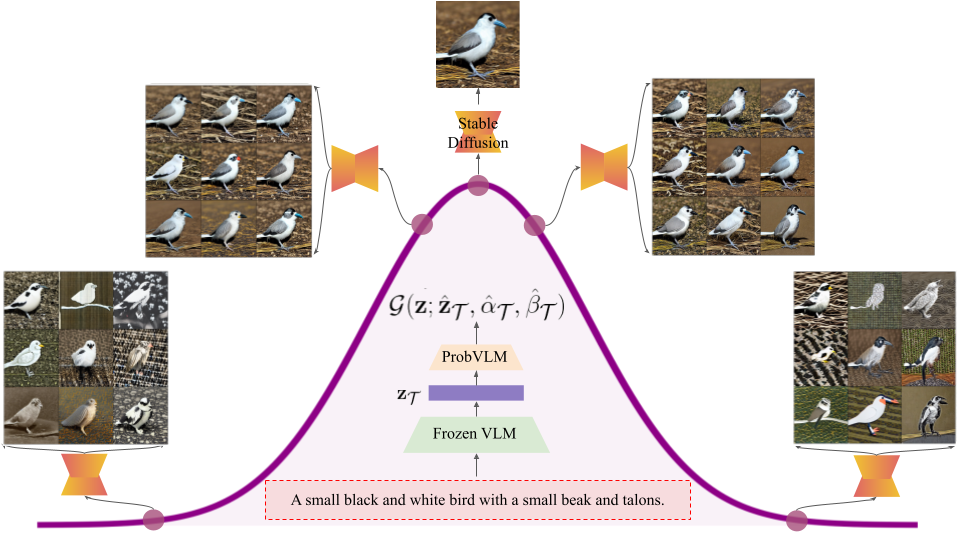}
     \vspace{-7pt}
     \caption{
     Visualizing the predicted embedding distributions from \texttt{ProbVLM} using a large-scale pre-trained diffusion model, i.e., \textit{Stable Diffusion}. 
     The example is shown for two different captions from CUB dataset, for which the point-estimate embedding vector is obtained via CLIP, and the distribution is obtained via \texttt{ProbVLM}.
     }
     \label{fig:stable-diff}
     \vspace{-10pt}
\end{figure*}

\myparagraph{Active Learning.} Here, we choose a small subset of the unlabeled dataset to fine-tune the model~\cite{svp}. In this case, we wish to finetune the CLIP model on the FLO dataset while using a limited amount of labeled data. 
To achieve this, we estimate the uncertainty of the image embeddings using \texttt{ProbVLM} (trained using a diverse dataset like COCO).
We then select the top-k samples sorted by their mean uncertainty in the visual embeddings and fine-tune the CLIP model using them with a contrastive objective~\cite{clip}. 
Results with varying budgets are shown in Figure~\ref{fig:active}. Selecting samples based on uncertainty scores significantly outperforms random sampling for all considered visual backbones.

\myparagraph{Pretrained Model Selection.}
We are given a set of models trained on different data distributions. We aim to select the best model for the target distribution for which we have unlabeled samples. This has been explored mostly in the context of classification previously~\cite{garg2022leveraging,guillory2021predicting,chen2021mandoline,chuang2020estimating,deng2021does,deng2021labels}. 
\begin{figure}[t]
{
\setlength{\tabcolsep}{5pt}
\renewcommand{\arraystretch}{1.2}
\resizebox{\linewidth}{!}{
\begin{tabular}{llcccc}
  \multirow{2}{*}{} & \multirow{2}{*}{} & \multicolumn{4}{c}{\textbf{Metrics}} \\
  \textbf{D} & \textbf{Models} & \textbf{Uncertainty score} & \textbf{R@1} & \textbf{R@5} & \textbf{R@10} \\
  \toprule
  \multirow{3}{*}{\rotatebox[origin=c]{90}{\textbf{CUB}}}  
  & CLIP-ViT32-COCO & 11.92 & 31.5 & 61.0 & 75.8  \\
  & CLIP-ViT32-Flickr & \textbf{9.37} & \textbf{32.4} & \textbf{64.2} & \textbf{76.9}  \\
  & CLIP-ViT32-FLO & 15.43 & 22.8 & 49.8 & 64.9  \\
  \midrule
  \multirow{3}{*}{\rotatebox[origin=c]{90}{\textbf{FLO}}}  
  & CLIP-ViT32-COCO & \textbf{11.83} & 47.9 & 79.2 & 88.5  \\
  & CLIP-ViT32-Flickr & 13.55 & \textbf{49.5} & \textbf{84.6} & \textbf{93.9}  \\
  & CLIP-ViT32-CUB & 18.39 & 37.7 &	69.4 &	82.8  \\
  \midrule
  \multirow{3}{*}{\rotatebox[origin=c]{90}{\textbf{Flickr}}}  
  & CLIP-ViT32-COCO & \textbf{9.61} & \textbf{88.8} & \textbf{97.8} & \textbf{99.8}\\
  & CLIP-ViT32-CUB & 16.49 & 25.8 & 47.4 & 55.6  \\
  & CLIP-ViT32-FLO & 13.67 & 52.8 & 77.8 & 85.2  \\
  \midrule
  \multirow{3}{*}{\rotatebox[origin=c]{90}{\textbf{COCO}}}  
  & CLIP-ViT32-Flickr & \textbf{7.28} & \textbf{58.1} & \textbf{80.9} & \textbf{88.2}  \\
  & CLIP-ViT32-CUB & 15.37 & 8.8	& 21.7 & 29.8  \\
  & CLIP-ViT32-FLO & 12.44 & 23.9 & 46.6 & 58.8  \\
  \midrule
\end{tabular}%
}
\vspace{-8pt}
\captionof{table}{
Results for the model selection experiment. \texttt{ProbVLM} accurately identifies the best performing source model using only unlabeled samples of the target dataset.
}
\label{tab:selection}
}
\vspace{-15pt}
\end{figure}

We consider the specific case of having the CLIP models fine-tuned on three datasets, and the fourth dataset is held out, for which we only have the images. We compute the mean uncertainty on these images using \texttt{ProbVLM} whose weights are interpolated from all the source datasets~\cite{soups,wortsman2022robust,patching,taskvectors}. This is to ensure that the uncertainties on all these models are comparable. The results for this experiment are shown in Table~\ref{tab:selection}. On CUB, Flickr, and COCO, the source model with the lowest uncertainty has the best performance on the target dataset, and on FLO dataset, the model with the least uncertainty has the 2nd best performance (R@1 of 47.9 vs 49.5 for the best model). 
This indicates that the uncertainties provided by \texttt{ProbVLM} can be used as a signal to predict the performance on unlabelled samples for retrieval.

\subsection{Latent Diffusion for Embedding Uncertainty }
\label{sec:diffusion}
To further understand the semantics of the predicted embedding distributions from the \texttt{ProbVLM}, 
we visualize the text embedding distributions by sampling the embedding vectors from the predicted distribution for a caption (converted to embedding vector using CLIP) and passing it through the pre-trained latent diffusion model using CLIPs text encoder, \textit{stable diffusion}, as shown in Figure~\ref{fig:stable-diff} and described in details in Section~\ref{sec:st}.
We observe from Figure~\ref{fig:stable-diff} that the samples obtained closer to the mean (i.e., sampled embedding vector similar to the one generated by CLIP for the caption) lead to meaningful variations in the generated images, e.g., for the left caption, close to the mean of the distribution, the generated samples show variations in the shape and colour of the beak, wings, and feet. Whereas far away from the mean of the distributions, i.e., on the tails, we start seeing images with strong artifacts that no longer preserves the semantics of the caption. We observe this for another example as well shown in Figure~\ref{fig:stable-diff}-(Right). More results are available in the supplementary.

\section{Conclusion}
\label{sec:conclusion}
We introduce \texttt{ProbVLM}, a post-hoc method for estimating the embedding distribution for a frozen large-scale deterministic vision-language model. 
We efficiently estimate calibrated uncertainties using our framework and show that such calibrated estimates have a variety of applications in downstream tasks such as model selection and active learning. 
Furthermore, we perform experiments to interpret embedding distribution predicted by \texttt{ProbVLM} using a large-scale pre-trained latent diffusion model (i.e., \textit{Stable Diffusion}). 
We hope our work highlights and inspires future work on efficient methods for probabilistic embeddings.

\myparagraph{Acknowledgements.} This work was supported by DFG project number 276693517, by BMBF FKZ: 01IS18039A, by the ERC (853489 - DEXIM), by EXC number 2064/1 – project number 390727645, and by the MUR PNRR project FAIR - Future AI Research (PE00000013) funded by NextGenerationEU. The authors thank the International Max Planck Research School for Intelligent Systems (IMPRS-IS) for supporting Uddeshya Upadhyay and Shyamgopal Karthik.

\clearpage
\section*{APPENDIX}
\appendix
\section{Additional Theoretical Support}
We discuss Equation 4 from the main paper and how we simplify the same to obtain a loss function suitable for training deep learning models.
Given an image and text embedding pair $(\mathbf{z}_\mathcal{V}, \mathbf{z}_\mathcal{T})$ (from frozen model) representing similar concepts, the output distributions from $\mathbf{\Psi}(\cdot; \zeta)$, 
 $\mathcal{G}(\mathbf{z}; \hat{\mathbf{z}}_{\mathcal{V}}, \hat{\mathbf{\alpha}}_{\mathcal{V}}, \hat{\mathbf{\beta}}_{\mathcal{V}})$ and $\mathcal{G}(\mathbf{z}; \hat{\mathbf{z}}_{\mathcal{T}}, \hat{\mathbf{\alpha}}_{\mathcal{T}}, \hat{\mathbf{\beta}}_{\mathcal{T}})$ (later referred to as $\mathcal{G}_{\mathcal{V}}(\mathbf{z})$) and $\mathcal{G}_{\mathcal{T}}(\mathbf{z})$) should match. 
 This can be measured directly from the likelihood as, $p(\mathbf{z}_v = \mathbf{z}_u)$, where 
$\mathbf{z}_v \sim \mathcal{G}_{\mathcal{V}}(\mathbf{z})$ and
$\mathbf{z}_u \sim \mathcal{G}_{\mathcal{T}}(\mathbf{z})$ as in~\cite{pfe} 
, i.e.,
\begin{gather}
    \scalebox{1.0}{$
    p(\mathbf{z}_v = \mathbf{z}_u)
    :=$} 
    \iint  
    \scalebox{0.99}{$
    \mathcal{G}_{\mathcal{V}}(\mathbf{z}_v) \mathcal{G}_{\mathcal{T}}(\mathbf{z}_u) \delta(\mathbf{z}_v-\mathbf{z}_u) d\mathbf{z}_v d\mathbf{z}_u
    \label{eq:eq1}
    $}
\end{gather}
where $\delta(\cdot)$ refers to the \textit{Dirac-delta distribution}.
The above integral can be simplified further by defining $\Delta \mathbf{z} = \mathbf{z}_v - \mathbf{z}_u$ and seeking $p(\Delta \mathbf{z}) = 0$. As both $\mathbf{z}_v$ and $\mathbf{z}_u$ are GGD random variables, $\Delta \mathbf{z}$ follows the distribution based on the \textit{Bivariate Fox H-function}~\cite{BFHF} given by,
\begin{gather}
    \Delta z \sim \scalebox{0.9}{$ \frac{1}{2 \Gamma(1/\hat{\mathbf{\beta}}_{\mathcal{V}}), \Gamma(1/\hat{\mathbf{\beta}}_{\mathcal{T}})} $} \times
     \nonumber \\
    \int
    \scalebox{0.8}{$
    \mathcal{H}^{1,1}_{1,2} \left[ A t^2 \big| 
 \begin{matrix}
 (1-\frac{1}{\hat{\mathbf{z}}_{\mathcal{V}}}, \frac{1}{\hat{\mathbf{z}}_{\mathcal{T}}}) \\
 (0,1) (\frac{1}{2}, 1)
 \end{matrix}
 \right]
    \mathcal{H}^{1,1}_{1,2} \left[ B t^2 \big| 
 \begin{matrix}
 (1-\frac{1}{\hat{\mathbf{z}}_{\mathcal{T}}}, \frac{1}{\hat{\mathbf{z}}_{\mathcal{T}}}) \\
 (0,1) (\frac{1}{2}, 1)
 \end{matrix}
 \right] \cos{t(\mu - z)}
    $}
    dt
    \label{eq:bfhf}
\end{gather}
Where $A=\frac{\hat{\mathbf{\alpha}}_{\mathcal{V}}^2 \Gamma(1/\hat{\mathbf{\beta}}_{\mathcal{V}})}{4 \Gamma(3/\hat{\mathbf{\beta}}_{\mathcal{V}})}$, $B=\frac{\hat{\mathbf{\alpha}}_{\mathcal{T}}^2 \Gamma(1/\hat{\mathbf{\beta}}_{\mathcal{T}})}{4 \Gamma(3/\hat{\mathbf{\beta}}_{\mathcal{T}})}$, $\mu=\hat{\mathbf{z}}_v - \hat{\mathbf{z}}_u$, and $\mathcal{H}$ is the \textit{Fox H function}~\cite{BFHF} given by,
\begin{gather}
H_{p, q}^{m, n}\left[z \mid \begin{array}{llll}
\left(a_1, A_1\right) & \left(a_2, A_2\right) & \ldots & \left(a_p, A_p\right) \\
\left(b_1, B_1\right) & \left(b_2, B_2\right) & \ldots & \left(b_q, B_q\right)
\end{array}\right] \nonumber \\
=
\frac{1}{2 \pi i} \int_L
\scalebox{0.93}{$
 \frac{\prod_{j=1}^m \Gamma\left(b_j+B_j s\right) \prod_{j=1}^n \Gamma\left(1-a_j-A_j s\right)}{\prod_{j=m+1}^q \Gamma\left(1-b_j-B_j s\right) \prod_{j=n+1}^p \Gamma\left(a_j+A_j s\right)} z^{-s} ds
$}
\end{gather}

Equation~\ref{eq:bfhf} does not provide a scalable objective function suitable for training deep neural networks. 
Hence, we propose an approximation that is easily scalable for deep-learning models given by,
\begin{gather}
    p(\mathbf{z}_v = \mathbf{z}_u) =
    \iint  
    \mathcal{G}_{\mathcal{V}}(\mathbf{z}_v)
    \mathcal{G}_{\mathcal{T}}(\mathbf{z}_u)
    \delta(\mathbf{z}_v-\mathbf{z}_u) d\mathbf{z}_v d\mathbf{z}_u \nonumber \\
    \approx \int 
    \frac{1}{2}
    \left(
    \mathcal{G}_{\mathcal{V}}(\mathbf{z}) \delta(\mathbf{z} - \mathbf{z}_{\mathcal{T}}) + 
    \mathcal{G}_{\mathcal{T}}(\mathbf{z}) \delta(\mathbf{z} - \mathbf{z}_{\mathcal{V}})
    \right)
    d\mathbf{z}
    \label{eq:cm_like}
\end{gather}

\begin{figure}[t]
    \centering
    \includegraphics[width=0.48\textwidth,page=1]{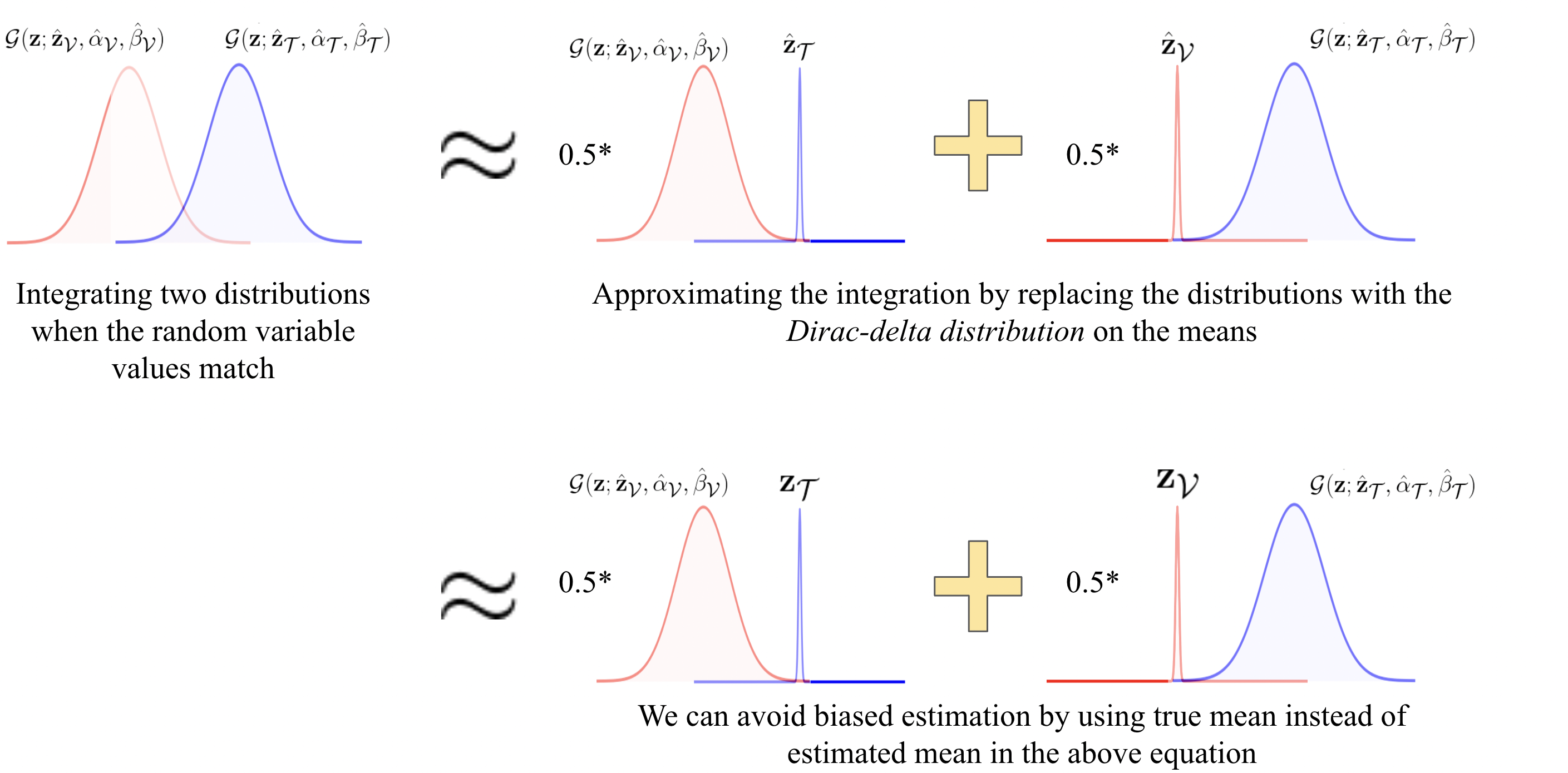}
    \caption{Visualizing the approximation in Equation~\ref{eq:cm_like}.}
    \label{fig:aprx}
\end{figure}
To understand the above approximation, we refer to Figure~\ref{fig:aprx}.
We notice that the integral in Equation~\ref{eq:eq1} tries to convolve the two distribution, with an additional constraint of those distributions being equal in value. While convolving the two generalized gaussian distributions is hard, Figure~\ref{fig:aprx} shows that a rough approximation for the same is to convolve a generalized gaussian distribution with the Dirac-delta distribution. Further, instead of using the estimated means from \texttt{ProbVLM} in the Dirac-delta distribution (that are to be near-perfect reconstructions of the embeddings obtained from the frozen network), we use the embeddings from the frozen encoders as shown in Figure~\ref{fig:aprx}.
This finally leads to Equation~\ref{eq:cm_like}.
The first term in the integral, $\int \mathcal{G}_{\mathcal{V}}(\mathbf{z}) \delta(\mathbf{z} - \mathbf{z}_{\mathcal{T}}) d\mathbf{z}$, is the likelihood of the text embedding $\mathbf{z}_{\mathcal{T}}$ under the predicted distribution, $\mathcal{G}_{\mathcal{V}}(\mathbf{z})$, for the visual embedding.
Similarly, the second term is the likelihood of the visual embedding $\mathbf{z}_{\mathcal{V}}$ under the predicted distribution, $\mathcal{G}_{\mathcal{T}}(\mathbf{z})$, for the text embedding. Negative log of Equation~\ref{eq:cm_like} leads to a scalable objective function that can be used to learn the optimal parameters for vision and text components of \texttt{ProbVLM} ($\mathbf{\Psi}_{\mathcal{V}}(\cdot; \zeta_{\mathcal{V}})$ and $\mathbf{\Psi}_{\mathcal{T}}(\cdot; \zeta_{\mathcal{T}})$),
\begin{gather}
    L_{\text{cross}}(\zeta_{\mathcal{V}}, \zeta_{\mathcal{T}}) :=  %
    \scalebox{0.9}{$
    \underbrace{
    \left( \frac{|\hat{\mathbf{z}}_{\mathcal{V}} - \mathbf{z}_{\mathcal{T}}|}{\hat{\mathbf{\alpha}}_{\mathcal{V}}} \right)^{\hat{\mathbf{\beta}}_{\mathcal{V}}} - \log \frac{\hat{\mathbf{\beta}}_{\mathcal{V}}}{\hat{\mathbf{\alpha}}_{\mathcal{V}}} + \log \Gamma(\frac{1}{\hat{\mathbf{\beta}}_{\mathcal{V}}})
    }_{\text{Cross-modal: vision} \rightarrow \text{text}}
    $}
    + \nonumber \\
    \scalebox{0.9}{$
    \underbrace{
    \left( \frac{|\hat{\mathbf{z}}_{\mathcal{T}} - \mathbf{z}_{\mathcal{V}}|}{\hat{\mathbf{\alpha}}_{\mathcal{T}}} \right)^{\hat{\mathbf{\beta}}_{\mathcal{T}}} - \log \frac{\hat{\mathbf{\beta}}_{\mathcal{T}}}{\hat{\mathbf{\alpha}}_{\mathcal{T}}} + \log \Gamma(\frac{1}{\hat{\mathbf{\beta}}_{\mathcal{T}}})
    }_{\text{Cross-modal: text} \rightarrow \text{vision}}
    $}
\end{gather}

In practice, the exponential of $\beta$ in the above equation often makes training unstable.
To make it more stable, we make use of the Taylor-series expansion and note that
\begin{gather}
    \left( \frac{|\hat{\mathbf{z}} - \mathbf{z}|}{\hat{\alpha}} \right)^{\hat{\beta}} = \left( 1 + (\frac{|\hat{\mathbf{z}} - \mathbf{z}|}{\hat{\alpha}} - 1) \right)^{\hat{\beta}} \nonumber \\ 
    \approx 1 - \hat{\beta} + \hat{\beta} (\frac{|\hat{\mathbf{z}} - \mathbf{z}|}{\hat{\alpha}})
\end{gather}
This way, the variable $\hat{\beta}$ no longer in exponent and as a result loss becomes more stable during optimization.

\begin{table}[t]
{
\setlength{\tabcolsep}{4pt}
\renewcommand{\arraystretch}{1.3}
\resizebox{\linewidth}{!}{
\begin{tabular}{llccc ccc ccc ccc}
  \multirow{3}{*}{} & \multirow{3}{*}{} & \multicolumn{12}{c}{\textbf{Datasets}} \\
  & & \multicolumn{3}{c}{\textbf{CUB}} & \multicolumn{3}{c}{\textbf{Flowers}} & \multicolumn{3}{c}{\textbf{Flickr}} & \multicolumn{3}{c}{\textbf{COCO}} \\
  \cmidrule(lr){3-5} \cmidrule(lr){6-8} \cmidrule(lr){9-11} \cmidrule(lr){12-14}
  \textbf{M} &  & R@1 & R@5 & R@10 & R@1 & R@5 & R@10 & R@1 & R@5 & R@10 & R@1 & R@5 & R@10 \\ 
  \toprule
  \multirow{6}{*}{\textbf{V-B32}}
  & \multirow{3}{*}{\textbf{i2t}}
  & 35.3 & 64.9 & 79.3 & 54.5 & 84.7 & 94.0 & 79.0 & 94.7 & 97.1 & 50.6 & 75.0 & 83.6 \\
  & & \textbf{85.1} & 89.4 & 81.9 & 53.3 & 55.2 & 37.2 & 64.2 & 61.0 & 55.1 & 61.0 & 62.3 & 57.2 \\
  & & \textbf{92.1} & 95.0 & 90.1 & 69.6 & 70.6 & 52.3 & 77.0 & 73.6 & 68.8 & 75.8 & 76.5 & 73.3 \\
  & \multirow{3}{*}{\textbf{t2i}} 
  & 16.0 & 34.4 & 44.6 & 25.5 & 47.8 & 61.8 & 56.5 & 82.2 & 88.3 & 30.1 & 55.7 & 66.8 \\
  & & \textbf{63.9} & 63.0 & 60.5 & 37.3 & 33.5 & 31.7 & 36.2 & 35.5 & 35.1 & 35.9 & 36.9 & 35.4 \\
  & & \textbf{72.8} & 71.8 & 70.7 & 47.4 & 43.3 & 43.7 & 47.5 & 46.9 & 46.7 & 47.2 & 49.3 & 47.8 \\
  \midrule
  \multirow{6}{*}{\textbf{V-B16}}
  & \multirow{3}{*}{\textbf{i2t}}
  & 34.2 & 66.2 & 80.4 & 52.1 & 82.8 & 91.6 & 82.7 & 96.2 & 98.9 & 53.0 & 77.1 & 85.1 \\
  & & \textbf{85.1} & 89.4 & 81.9 & 53.3 & 55.2 & 37.2 & 64.2 & 61.0 & 55.1 & 61.0 & 62.3 & 57.2 \\
  & & \textbf{92.1} & 95.0 & 90.1 & 69.6 & 70.6 & 52.3 & 77.0 & 73.6 & 68.8 & 75.8 & 76.5 & 73.3 \\
  & \multirow{3}{*}{\textbf{t2i}}
  & 15.0 & 33.3 & 44.1 & 25.4 & 46.4 & 57.9 & 61.0 & 84.2 & 89.6 & 33.3 & 58.6 & 68.9 \\
  & & \textbf{63.9} & 63.0 & 60.5 & 37.3 & 33.5 & 31.7 & 36.2 & 35.5 & 35.1 & 35.9 & 36.9 & 35.4 \\
  & & \textbf{72.8} & 71.8 & 70.7 & 47.4 & 43.3 & 43.7 & 47.5 & 46.9 & 46.7 & 47.2 & 49.3 & 47.8 \\
  \midrule
  \multirow{6}{*}{\textbf{RN-50}}
  & \multirow{3}{*}{\textbf{i2t}}
  & 31.1 & 61.7 & 75.9 & 53.0 & 87.1 & 95.0 & 77.7 & 95.2 & 97.3 & 49.1 & 72.5 & 81.8 \\
  & & \textbf{85.1} & 89.4 & 81.9 & 53.3 & 55.2 & 37.2 & 64.2 & 61.0 & 55.1 & 61.0 & 62.3 & 57.2 \\
  & & \textbf{92.1} & 95.0 & 90.1 & 69.6 & 70.6 & 52.3 & 77.0 & 73.6 & 68.8 & 75.8 & 76.5 & 73.3 \\
  & \multirow{3}{*}{\textbf{t2i}}
  & 15.3 & 35.0 & 46.5 & 31.5 & 54.3 & 66.7 & 55.1 & 81.2 & 87.9 & 28.3 & 53.1 & 64.3 \\
  & & \textbf{63.9} & 63.0 & 60.5 & 37.3 & 33.5 & 31.7 & 36.2 & 35.5 & 35.1 & 35.9 & 36.9 & 35.4 \\
  & & \textbf{72.8} & 71.8 & 70.7 & 47.4 & 43.3 & 43.7 & 47.5 & 46.9 & 46.7 & 47.2 & 49.3 & 47.8 \\
  \midrule
  
\end{tabular}%
}
\vspace{-5pt}
\captionof{table}{
Zero-shot performance on COCO, Flickr, CUB and FLO with for both Image-to-Text (i2t) and Text-to-Image (t2i) Retrieval for CLIP Models (M) with Vision Transformer (V-B32, V-B16) and ResNet (RN-50) backbones. 
}
\label{tab:zs-clip}
}
\end{table}

\begin{table}[t]
{
\setlength{\tabcolsep}{3pt}
\renewcommand{\arraystretch}{2.2}
\resizebox{\linewidth}{!}{
\begin{tabular}{ll  ccc ccc ccc ccc}
  \multirow{3}{*}{} & \multirow{3}{*}{} & \multicolumn{12}{c}{\textbf{CLIP backbones fine-tuned on}} \\
  & & \multicolumn{3}{c}{\textbf{CUB}} & \multicolumn{3}{c}{\textbf{Flowers}} & \multicolumn{3}{c}{\textbf{Flickr}} & \multicolumn{3}{c}{\textbf{COCO}} \\
  \cmidrule(lr){3-5} \cmidrule(lr){6-8} \cmidrule(lr){9-11} \cmidrule(lr){12-14}
  \textbf{D} &  & V-B32 & V-B16 & RN-50 & V-B32 & V-B16 & RN-50 & V-B32 & V-B16 & RN-50 & V-B32 & V-B16 & RN-50 \\ 
  \toprule
  \multirow{2}{*}{\rotatebox[origin=c]{90}{\textbf{CUB}}}
  & \multirow{1}{*}{\textbf{i2t}}
  & \textbf{58.8} & \textbf{66.1} & \textbf{53.9} & 25.2 & 23.8 & 13.4 & 32.4 & 31.1 & 26.2 & 31.5 & 32.5 & 26.8 \\
  & \multirow{1}{*}{\textbf{t2i}} 
  & \textbf{41.3} & \textbf{42.3} & \textbf{37.4} & 18.4 & 16.8 & 13.1 & 16.6 & 17.1 & 16.1 & 16.6 & 16.9 & 14.3 \\
  \midrule
  \multirow{2}{*}{\rotatebox[origin=c]{90}{\textbf{Flowers}}}
  & \multirow{1}{*}{\textbf{i2t}}
  & 54.5 & 51.1 & 44.3 & \textbf{80.7} & \textbf{82.0} & \textbf{73.8} & 49.5 & 55.2 & 49.7 & 47.9& 47.2 & 43.6 \\
  & \multirow{1}{*}{\textbf{t2i}} 
  & 25.5 & 31.2 & 29.6 & \textbf{57.8}  & \textbf{59.0} & \textbf{53.3} & 31.3 & 29.3 & 30.8 & 28.7 & 29.2 & 31.7\\
  \midrule
  \multirow{2}{*}{\rotatebox[origin=c]{90}{\textbf{Flickr}}}
  & \multirow{1}{*}{\textbf{i2t}} 
  & 68.9 & 73.5 & 48.2 & 51.4  & 62.4 & 24.4 & \textbf{90.0} & \textbf{92.7} & \textbf{87.1} & 86.7 & 90.2 & 87.7\\
  & \multirow{1}{*}{\textbf{t2i}} 
  & 48.6 & 54.7 & 31.4 & 32.3  & 40.5 & 17.0 & \textbf{73.4} & \textbf{77.5} & \textbf{68.3} & 69.9 & 74.5 & 68.7\\
  \midrule
  \multirow{2}{*}{\rotatebox[origin=c]{90}{\textbf{COCO}}}
  & \multirow{1}{*}{\textbf{i2t}} 
  & 32.6 & 42.6 & 22.0 & 24.8  & 31.8 & 8.9 & 56.9 & 61.5 & 52.0 & \textbf{73.4} & \textbf{69.5} & \textbf{64.3}\\
  & \multirow{1}{*}{\textbf{t2i}} 
  & 19.5 & 27.1 & 12.5 & 32.3  & 19.7 & 6.8 & 38.7 & 43.9 & 33.0 & \textbf{49.8}& \textbf{52.3} & \textbf{45.3} \\
  \midrule
\end{tabular}%
}
\vspace{-10pt}
\captionof{table}{
Result for fine-tuning CLIP on different Datasets (D) for Image-to-Text (i2t) and Text-to-Image (t2i) retrieval.
}
\label{tab:ft-clip}
}
\end{table}

\begin{figure}[h]
    \centering
    \includegraphics[width=0.49\textwidth]{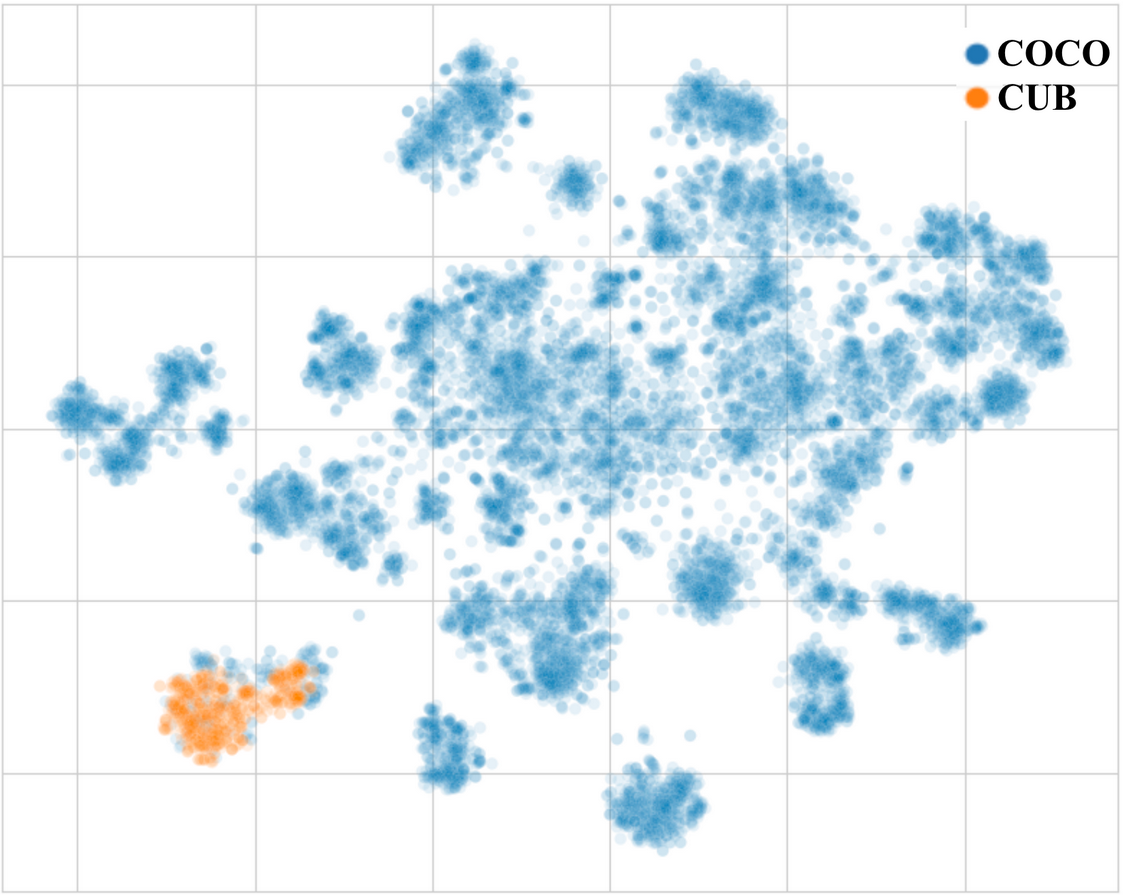}
    \vspace{-10pt}
    \caption{tSNE plot for MS-COCO and CUB image embeddings illustrating the diversity of MS-COCO.}
    \label{fig:tsne}
\end{figure}

\section{Additional Quantitative Experiments}
We provide the zero-shot results for the CLIP model trained with different visual backbones in Table~\ref{tab:zs-clip}, while the results after fine-tuning are presented in Table~\ref{tab:ft-clip}. While Zero-Shot CLIP achieves promising results on all 4 datasets, these are much worse when compared to the results obtained when fine-tuning on the desired target dataset (42.3 vs. 15.0 for a ViT B/16 on CUB t2i R@1). However, this comes at the cost of worse performance on the remaining datasets due to catastrophic forgetting and has to be mitigated via several strategies.

Figure~\ref{fig:tsne} shows the tSNE plots for the CLIP embeddings obtained from a relatively diverse dataset (e.g., COCO) compared to a niche dataset (e.g., CUB consisting of only birds). 
As indicated in the plot, a niche dataset will likely not be able to capture all the representations spread in the embedding space, leading to poor generalization, as shown in Table~\ref{tab:ft-clip}. This is because CUB has images that only contain birds, whereas COCO is a much larger datasets containing 80 different object categories (including birds). Therefore fine-tuning either the VLM or \texttt{ProbVLM} on a larger, more diverse dataset such as COCO would lead to better generalization, and trasnferrability across datasets.

\section{Implementation Details and Code}
\myparagraph{Stable Diffusion interpretation.} The U-Net based decoder used by Stable Diffusion takes the pre-final layer of the CLIP text encoder as input, which expects the input to be in a shape  \textit{tokens} x \textit{features} %
However, the usual training of \texttt{ProbVLM} takes the pooled output from the text encoder to enable cross-modal alignment with the vision encoder. For this experiment, we re-train \texttt{ProbVLM} to operate on the pre-final layer without the cross-modal alignment.

{\small
\bibliographystyle{ieee_fullname}
\bibliography{egbib}
}

\end{document}